%% file: main.tex
\documentclass[10pt,twocolumn,letterpaper]{article}

\usepackage{cvpr}
\input{preamble}

\definecolor{cvprblue}{rgb}{0.21,0.49,0.74}
\usepackage[pagebackref,breaklinks,colorlinks,allcolors=cvprblue]{hyperref}

\def\confName{CVPR}
\def\confYear{2027}

\makeatletter
\patchcmd{\@maketitle}{\vspace*{24pt}}{\vspace*{14pt}}{}{}
\patchcmd{\abstract}{\vspace*{12pt}\noindent}{\vskip 4pt\noindent}{}{}
\patchcmd{\endabstract}{\vspace*{12pt}}{\par}{}{}
\makeatother

\title{\texorpdfstring{\raisebox{-1.6pt}{\includegraphics[height=15.5pt]{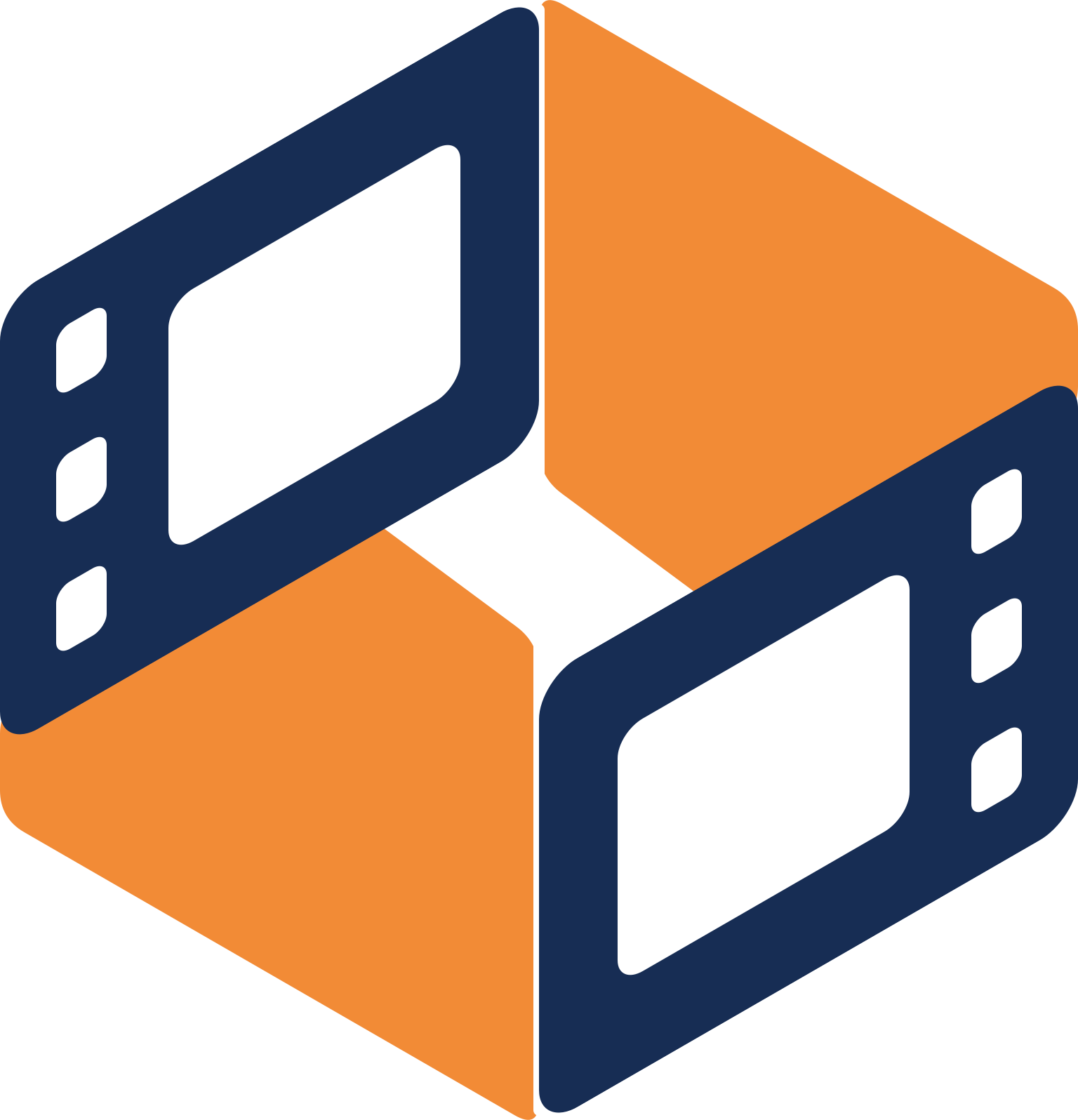}}\hspace{0.32em}}{}StateWeaver: Multi-Agent Coordination of Story-State Transitions for Long-Horizon Video Generation}

\author{
  Modi Jin\textsuperscript{1*}\quad
  Xuying Zhang\textsuperscript{2,1}\quad
  Yijing Guo\textsuperscript{3}\quad
  Haoran Li\textsuperscript{2}\quad
  Yanwen Ma\textsuperscript{4}\quad
  Jing Wang\textsuperscript{1}\quad
  Chenyu Ge\textsuperscript{1}\\[3pt]
  Ming-Ming Cheng\textsuperscript{1}\quad
  Haoyang Huang\textsuperscript{2}\\[5pt]
  \textsuperscript{1}Nankai University\quad \textsuperscript{2}Joy Future Academy, JD\quad \textsuperscript{3}ShanghaiTech University \quad \textsuperscript{4}Beihang University
}

\AtBeginDocument{\hypersetup{pdfauthor={Modi Jin, Xuying Zhang, Yijing Guo, Haoran Li, Yanwen Ma, Jing Wang, Chenyu Ge, Ming-Ming Cheng, Haoyang Huang}}}

\begin{document}
\twocolumn[{%
  \renewcommand\twocolumn[1][]{#1}%
  \maketitle
  \vspace{-0.78cm}
  {\centering\large Project page: \url{https://ghost233lism.github.io/StateWeaver-page/}\par}
  \begin{center}
    \centering
    \setlength{\abovecaptionskip}{2pt}
    \includegraphics[width=\linewidth]{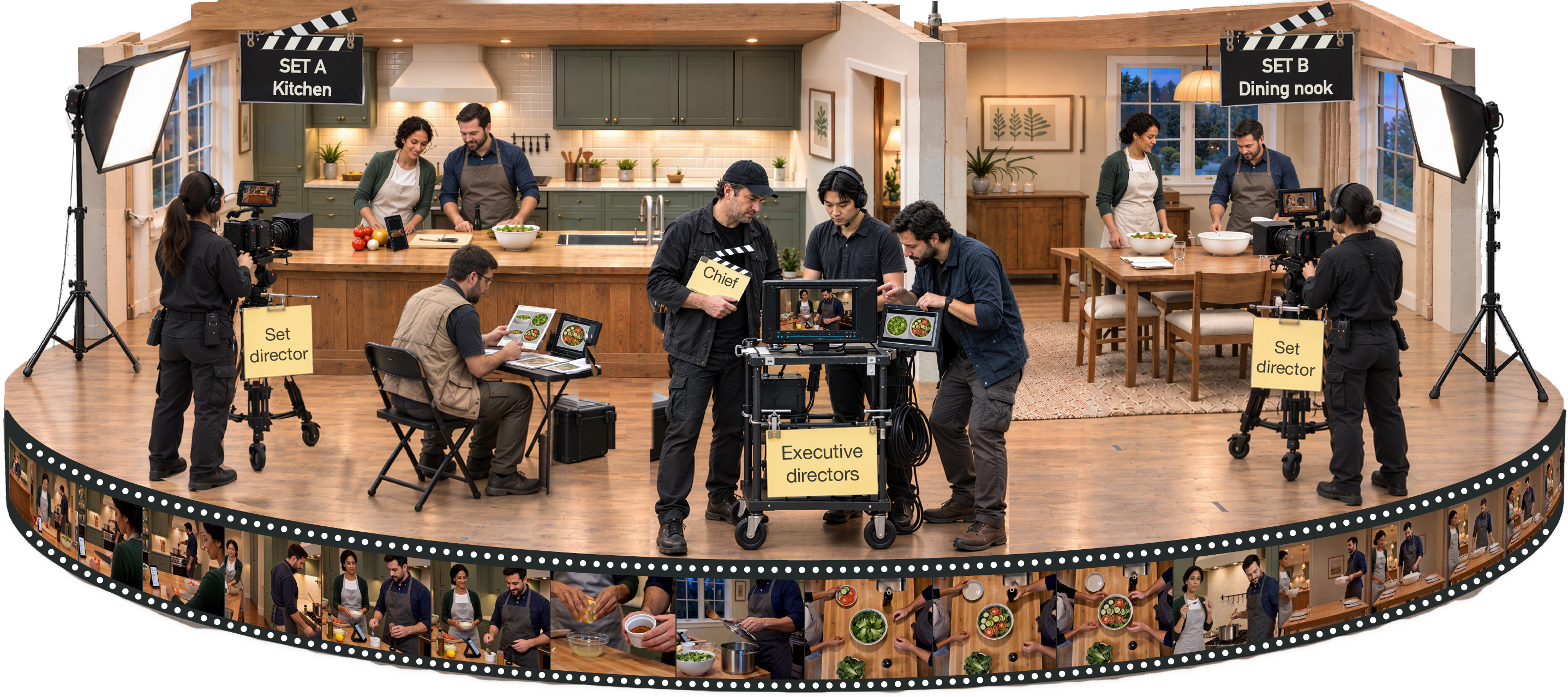}%
    \captionof{figure}{\textbf{\method enables coherent long-form video production through collaboration across sets.}
    Each set autonomously plans, generates, and repairs its sub-story, using validated visual evidence to maintain continuity across sets.
    }
    \label{fig:teaser}
  \end{center}%
}]

\enlargethispage{2\baselineskip}
\makeatletter
\begingroup
  \renewcommand{\thefootnote}{*}
  \renewcommand{\@makefntext}[1]{\noindent\@makefnmark\,#1}
  \footnotetext[0]{\textit{Work done during an internship at Joy Future Academy, JD.}}
\endgroup
\makeatother

\begin{abstract}
Modern video foundation models generate high-quality short clips, but preserving long-video consistency and correctly depicting event-driven state changes remain challenging.
Existing agentic workflows face these challenges because continual planning and repair accumulate context, while reference updates remain limited.
We introduce \method, which combines Active States and Hierarchical Collaboration to connect review-driven visual reference updates with cross-shot state inheritance.
Active States translate state discrepancies identified during review into concrete visual asset requirements that guide reference construction, validation, and regeneration, preserving identity and already-correct content while repairing target attributes.
Confirmed generation results are linked to the corresponding asset states, story time, and reference versions for verification and reuse by subsequent shots.
Hierarchical Collaboration isolates execution contexts by task, clip, and repair round, and uses hierarchical memory to organize this process across multiple sets.
Across multiple benchmarks, \method outperforms existing methods and produces long videos with sustained cross-shot consistency under predefined and autonomously planned storyboards.
\end{abstract}

\section{Introduction}
\label{sec:intro}

Rapid advances in video foundation models have made high-quality short-clip generation increasingly mature~\cite{gao2025seedance,wan2025wan,minimax2026h3}, while filmmaking, short-drama production, and embodied simulation demand long-horizon videos~\cite{duan2026joyaiecho15}.
Existing production platforms organize clip generation through editable workflows~\cite{runway2026workflows,ltx2026studio}, but provide limited support for using generation feedback to coordinate corrections across shots.
Local edits may therefore fail to preserve state continuity in subsequent clips and overall video coherence.
Native long-video models face challenges in preserving quality and controlling error accumulation over extended generation horizons~\cite{zhang2025framecontextpacking,yang2025longlive,cui2025selfforcingpp,yuan2026helios}.
Using short-clip foundation models as renderers, with agents autonomously coordinating planning, generation, and repair, offers a promising route to combining visual quality, narrative continuity, and production automation~\cite{yuan2024mora,wang2024aesopagent,xu2025mmstoryagent}.

\begin{figure}[t]
  \centering
 \includegraphics[width=0.95\linewidth]{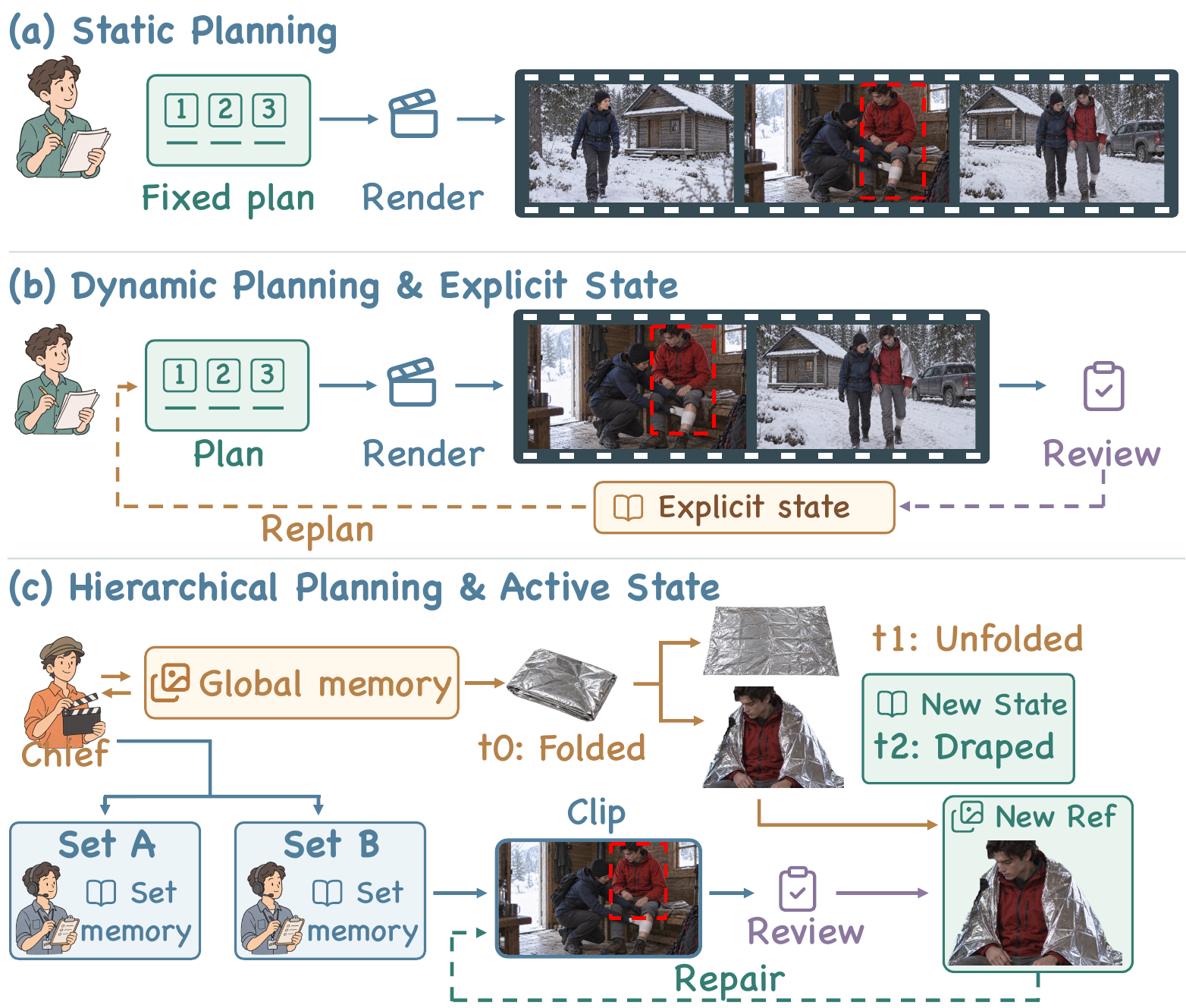}%
\caption{\textbf{Three paradigms of agentic long-video production.}}
  \label{fig:comparison}
\end{figure}

As shown in~\cref{fig:comparison}, early agentic long-video frameworks use specialized roles~\cite{zheng2024videogen,xu2025filmagent,wang2025mavis} for planning~\cite{he2023animateastory,huang2023freebloom,lin2023videodirectorgpt}, asset generation, and assembly, but their predefined plans~\cite{xie2024dreamfactory,hu2024storyagent,wu2025movieagent} can allow uncorrected generation errors to propagate into subsequent shots.
With advances in multimodal large language models~\cite{soni2024videoagent,huang2024genmac,long2025vista,song2026codirector}, MUSE~\cite{sun2026muse} and CineCrew~\cite{chen2026cinecrew} incorporate review feedback into production decisions~\cite{wei2025hollywoodtown,li2026frameworkers,liang2025univa}.
StoryEngine~\cite{wang2026storyengine} and ATAS~\cite{wang2026atas} further organize event-induced asset changes and validated observations through explicit states and temporal memory.
These advances support dynamic planning~\cite{wei2026videoweaver,liu2026cineforge}, state-based collaboration~\cite{zuo2026filmworld,kong2026mvagent,huang2026dramaagent}, and historical state reuse.
However, continual production across multiple sub-stories requires coordinating task-specific execution contexts and visual conditions for new states.

Specifically, explicitly recording target states before and after an event does not ensure that a video realizes the intended change.
Continual production needs to connect three steps.
First, review findings must identify the visual conditions missing from current references and guide the construction of targeted references.
Second, the state realized after regeneration must be confirmed so that subsequent shots inherit the repair instead of continuing to use outdated visual conditions.
Finally, as these operations recur across tasks and sub-stories, local execution contexts need to be isolated while sharing state evidence relevant to the current story interval~\cite{liu2026reca}.
Local visual repair and subsequent coherent production thus need to be addressed jointly.

\begin{figure*}[t]
  \centering
  \includegraphics[width=\linewidth]{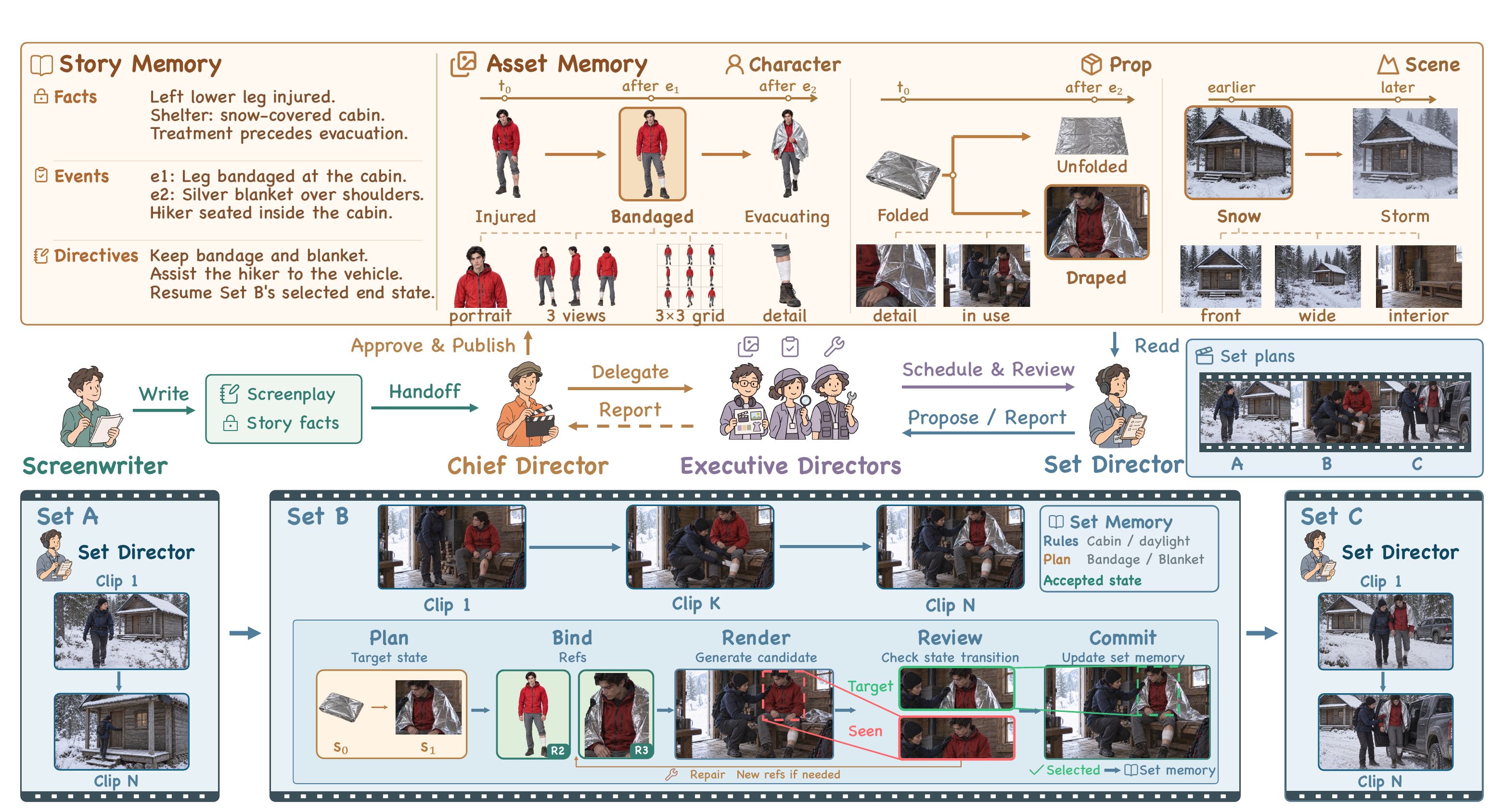}
  \caption{\textbf{Overview of \method.}
Active States support state planning, review-driven refinement, and reference supplementation, with confirmed results updating memory to preserve continuity across sets.
Hierarchical Collaboration organizes the chief director, executive directors, and set directors, connecting isolated execution contexts through hierarchical story and asset memory.
  }
  \label{fig:overview}
\end{figure*}

We therefore introduce \method (as shown in \cref{fig:teaser}), a multi-agent framework for continual long-video production that connects review-driven visual updates with the inheritance of repaired states.
Building on existing explicit state representations, we propose Active States to plan asset changes from established story events and translate discrepancies identified in actual generation and review into concrete visual asset requirements.
The system uses these requirements to construct and validate new visual conditions for targeted regeneration, preserving asset identity and already-correct content while repairing target attributes.
Confirmed generation results are linked to the corresponding asset states, story time, and reference versions and update hierarchical memory, enabling subsequent shots to verify and reuse the changes realized through repair.

Inspired by real-world filmmaking, we use Hierarchical Collaboration to organize continuous sub-stories as independent sets and isolate execution contexts by specialized task, clip planning, and repair round.
Hierarchical memory persistently stores production records, allowing agents to recover relevant plans, references, and confirmed observations when switching tasks.
Sets coordinate state handoffs through evidence dependencies.
This organization connects the information needed for local repair with shared production information across sets, supporting autonomous iteration within each set and parallel execution of independent tasks.

We conduct comparative evaluations on multiple video-generation benchmarks~\cite{zhang2025storymem,huang2026vimax,he2026entitybench,chen2026directorbench}.
The results show that \method outperforms existing frameworks and generates long videos that maintain cross-shot consistency under both predefined and autonomously planned storyboards.
Component ablations further demonstrate the contributions of state refinement and reference supplementation to state transitions and of context isolation to continual production.
Our contributions are summarized as follows.

\begin{itemize}[leftmargin=1.2em,itemsep=1pt,topsep=2pt]
  \item
  We propose Active States to translate discrepancies found during review into state-refinement and visual-reference supplementation requirements, linking confirmed outputs to asset states and story time for targeted repair.

  \item
  We propose Hierarchical Collaboration to isolate execution contexts by task, clip, and repair round, retrieve relevant information through hierarchical memory, and coordinate evidence-based state handoffs across sets for continual planning, local repair, and parallel production.

  \item
  Results on four benchmarks show that \method improves long-video consistency under both predefined and autonomously planned storyboards, with human evaluation and ablations further validating the core mechanisms.
\end{itemize}

\section{Related Work}
\label{sec:related}

\myPara{Video Generation.}
Modern video foundation models such as Seedance~\cite{gao2025seedance}, Wan~\cite{wan2025wan}, and MiniMax H3~\cite{minimax2026h3} improve generation through large-scale video training and can produce short clips with high visual fidelity and motion quality.
Early temporal co-denoising and streaming memory provide reusable mechanisms for extending clip generation~\cite{wang2023genlvideo,henschel2024streamingt2v}.
Coherent long videos, however, require contextual state to propagate across generation windows through temporal extension mechanisms~\cite{qiu2023freenoise,kim2024fifodiffusion,lu2024freelong,tan2024videoinfinity,zhao2025riflex} and explicit context management~\cite{yi2025deepforcing,li2026rollingsink,cai2025mixtureofcontexts}.
Recent methods learn sustained generation through hierarchical generation and transition modeling~\cite{yin2023nuwaxl,chen2023seine,zhao2024moviedreamer}, long-context conditioning~\cite{guo2025longcontexttuning,wang2024loong}, and autoregressive rollout~\cite{chen2025skyreelsv2,sand2025magi1,huang2025selfforcing}.
JoyAI-Echo-1.5~\cite{duan2026joyaiecho15} couples cross-shot audio-visual memory with rollout-aware training for persistent stories and interactive worlds.
Long rollouts still degrade continuity and short-clip quality~\cite{yang2025longlive,cui2025selfforcingpp}.

\myPara{Long-form and Agentic Video Generation.}
Agentic long-video frameworks organize short-clip generation into planning-to-assembly workflows~\cite{xie2024dreamfactory,hu2024storyagent}, drawing on story visualization~\cite{li2018storygan,pan2022coherentstory,rahman2022makeastory,gong2023talecrafter,liu2023intelligentgrimm} and storyboard planning~\cite{long2024videostudio,zhuang2024vlogger}.
Reference conditioning~\cite{huang2024incontextlora,wang2025storyanchors,zhang2026stage,vandersanden2026emvid} and subject consistency control~\cite{wang2023autostory,zhou2024storydiffusion,yang2024seedstory,he2024dreamstory,mao2024storyiter} preserve entity appearance, while multi-shot models~\cite{qi2025mask2dit,kara2025shotadapter,wu2025cinetrans,wang2025multishotmaster,meng2025holocine} and visual memory~\cite{zhou2026videomemory,an2025onestory,zhang2025storymem,ji2025memflow,huang2026unityshots,li2026joyaiecho} provide reusable cross-shot context.
Among the methods compared in this work, MovieAgent~\cite{wu2025movieagent} uses staged planning and assembly, ViMax~\cite{huang2026vimax} follows a fixed planning workflow, and StoryEngine~\cite{wang2026storyengine} and ATAS~\cite{wang2026atas} incorporate explicit states into generation.
Related work extends staged and fixed-plan production~\cite{huang2025filmaster,zhang2025anime,xiao2025captaincinema,mu2026scriptagent} through multi-candidate selection~\cite{shi2025animaker}, iterative feedback~\cite{sun2026muse,li2026videogenagent,zhang2026cineevolve}, and adaptive planning~\cite{chen2026cinecrew}.
Further advances include compact context allocation~\cite{liu2026reca}, verification of extracted visual memories~\cite{lai2026groundshot}, persistent state tracking~\cite{huang2026dramaagent,yin2026stateflow}, inference of subsequent entity states~\cite{miao2026stateagent,zhou2026phyevolve,bie2026evoworld}, and targeted visual repair~\cite{gao2026aesr,guan2026physrepair}.
However, sustained production still faces two connected challenges: retaining relevant context across repeated planning and repair, and reliably propagating corrected visual observations across shots.
Successful local repair alone does not ensure correct downstream reuse, which also depends on matching story intervals, asset states, source versions, and the scope under which each observation was verified.

\begin{algorithm}[t]
  \renewcommand{\rmdefault}{cmr}
  \renewcommand{\ttdefault}{cmtt}
  \SetAlFnt{\small\normalfont}
  \SetAlCapFnt{\small\fontfamily{cmr}\selectfont\bfseries}
  \SetAlCapNameFnt{\small\normalfont}
  \SetNlSty{textnormal}{}{}
  \SetAlgoNlRelativeSize{-2}
  \SetArgSty{textit}
  \SetFuncSty{textnormal}
  \SetCommentSty{texttt}
  \SetInd{0.5em}{0.8em}
  \SetKwInput{KwIn}{Input}
  \SetKwInput{KwOut}{Output}
  \SetKw{Return}{return}
  \DontPrintSemicolon
  \SetAlgoNoEnd
  \caption{\method production workflow.}
  \label{alg:production}
  \KwIn{Story request $u$}
  \KwOut{Long video $V$}
  $\Pi\gets\operatorname{Write}(u)$ and $\mathcal E_{\mathrm{acc}}\gets\varnothing$\;
  $(\{S_k\},M_G,\{M_{S_k}\},\mathcal G)\gets\operatorname{Initialize}(\Pi)$\;
  \While{unfinished tasks remain}{
    Prepare independent assets and shots\;
    $\mathcal Q\gets\operatorname{Ready}(\mathcal G,\mathcal E_{\mathrm{acc}})$\;
    \If{$\mathcal Q=\varnothing$ and no tasks are running}{
      AwaitChiefIntervention($\mathcal G$), then continue\;
    }
    Dispatch(Bind($\mathcal Q$)) \tcp*[r]{isolated contexts}
    $e\gets\operatorname{AwaitResult}$\;
    \If{$e$ is a candidate $V_i^{(t)}$}{
      $(\rho_i,D_i^{(t)},P_i^{(t)},N_i^{(t)})\gets\operatorname{Review}(V_i^{(t)})$\;
      $d\gets\mathrm{select}$ if $\rho_i=\mathrm{pass}$, else $\mathrm{repair}$\;
      \If{$d=\mathrm{repair}$ and budget exhausted}{
        $d\gets\operatorname{ChiefDecision}(i)$\;
      }
      \eIf{$d=\mathrm{select}$}{
        $\widehat V_i\gets\operatorname{Select}(i)$ and Record($\widehat V_i,M_{S_k}$)\;
        Publish(ChiefConfirm($\widehat V_i$), $M_G,\mathcal E_{\mathrm{acc}}$)\;
      }{
        \If{$N_i^{(t)}\ne\varnothing$ or $d=\mathrm{extra}$}{
          DelegateSupply($i,D_i^{(t)},P_i^{(t)},N_i^{(t)}$)\;
        }
        QueueRepair($i,D_i^{(t)},P_i^{(t)}$)\;
      }
    }
    Refresh($\mathcal G$) and Release($e$)\;
  }
  \Return Assemble(selected clips, playback order)\;
\end{algorithm}

\section{StateWeaver}
\label{sec:method}

\subsection{Framework Overview}
\label{sec:overview}

\cref{fig:overview,alg:production} summarize \method's production workflow.
The screenwriter expands a brief request into a screenplay specifying characters, events, and causal relations.
The chief director partitions temporally and spatially continuous sub-stories into sets, each with an event scope and production objectives.
Set directors plan, generate, review, and repair clips, while executive directors handle delegated asset preparation, repair diagnosis, candidate reassessment, and blocked tasks.
The chief director coordinates sets and confirms shared evidence.
When no task is ready, production waits for running tasks or, if none are running, for chief-directed intervention before scheduling resumes.
Active States support state planning, reference refinement, and local repair. Hierarchical Collaboration sustains these operations across sets through isolated contexts, hierarchical memory, and evidence dependencies.

\subsection{Active States}
\label{sec:state}

\myPara{State Trees and Story Time.}
Active States enable agents to plan, refine, and supplement asset states and visual references while maintaining consistent story facts.
\method organizes events and their story time from the screenplay, recording them separately from playback order and generation or repair rounds.
The chief director first specifies base asset states and set entry and exit requirements, then establishes initial plans for states with concrete reference purposes and refines or supplements states and references according to shot requirements and review feedback.
Each state is associated with its applicable story interval to support accurate reference use in flashbacks, cross-set revisits, and parallel production.
A state tree records the origins of an asset's changes, with each node represented as
\begin{equation}
  s=(a,\tau,\mathbf x,p,\mathcal E),
  \label{eq:state}
\end{equation}
Here, $a$ is the stable asset identity, $\tau$ is the applicable story interval, $\mathbf x$ contains mutable attributes such as appearance, location, and ownership, and $p$ is the principal parent state of the same asset, supported by story evidence $\mathcal E$.
Successor states retain unchanged parent attributes that remain applicable and update attributes affected by the current event.
Nodes sharing a parent record changes derived from different events, and their reference use is determined by their respective event scopes and story intervals.
Image generation or editing within a state produces reference revisions, with image provenance recorded separately.
Clips spanning state changes select entry and target states by event scope and story interval, and bind auxiliary references to exact versions.
Target states and references describe intended changes, while actual observations describe verifiable attributes in generated results.
Target states provide criteria for judging whether generated results meet requirements, but do not constitute evidence that those states have been realized.
Discrepancies in generated results must not alter established story requirements.
Images independent of upstream observations can be prepared in advance, whereas references that inherit actual upstream appearance are constructed once the corresponding observations are approved.
If a state is missing or its applicable scope conflicts with another, the chief director supplements the state or checks the event scope.
Affected clips wait for resolution while other sets and preparation tasks continue.

\begin{figure*}[t]
  \centering
    \includegraphics[width=\linewidth]{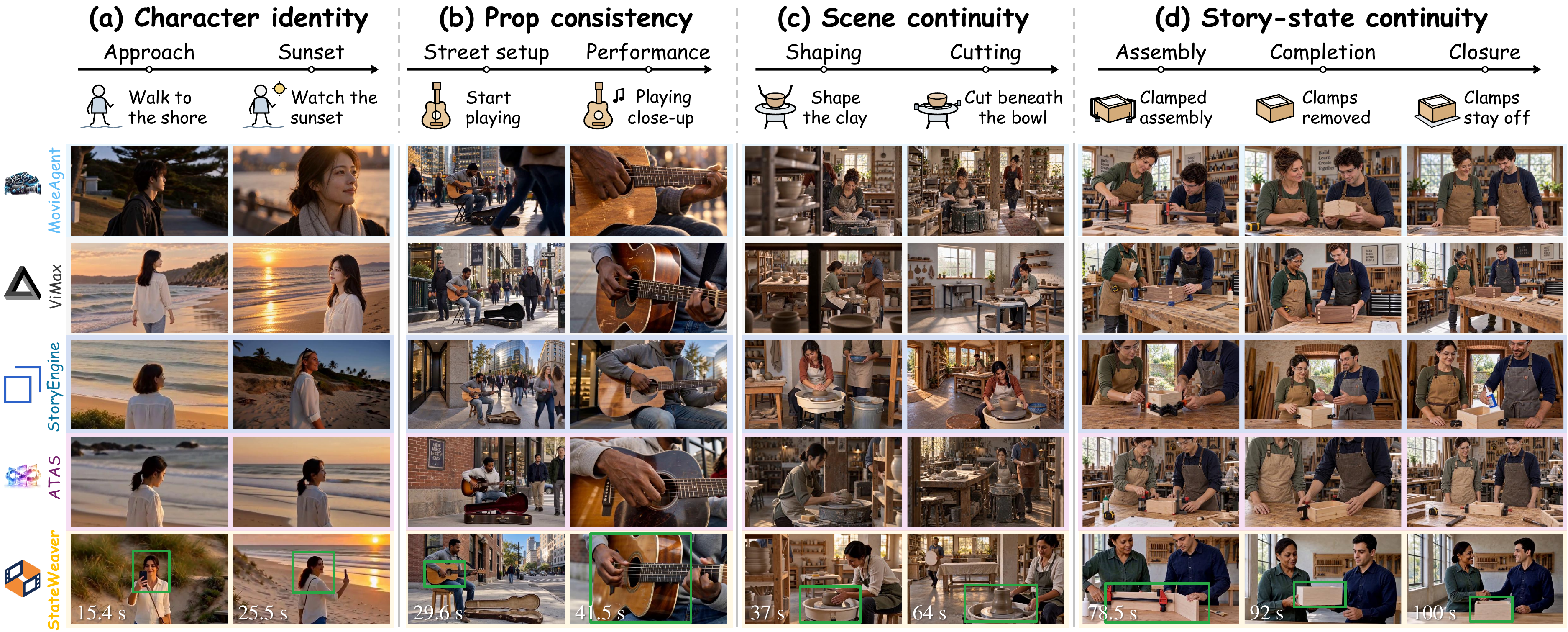}%
  \caption{\textbf{Long-range qualitative comparison.}
  Rows show MovieAgent, ViMax, StoryEngine, ATAS, and \method at corresponding story moments.
  \method preserves visual identity across shots and carries event-induced state changes into subsequent clips.}
  \label{fig:qualitative}
\end{figure*}

\myPara{State and Reference Refinement}
\method adjusts generation conditions through state-transition review, refining states and supplementing references as needed for local repair.
The set director checks identity, key actions, and their visual consequences against event requirements, entry and target states, and actual references, organizing round-$t$ discrepancies into defects $D_i^{(t)}$, content to preserve $P_i^{(t)}$, and asset requirements $N_i^{(t)}$.
Asset requirements link clip and state plans and specify target assets, story intervals, attributes to change, preservation constraints, reference purposes, and source evidence.
When existing references suffice, the candidate is regenerated directly. Missing materials prompt a request for reference expansion (RE) to the chief director.
RE constructs missing states or adds and revises references for views, actions, layouts, and target frames within an existing state.
Defect-conditioned reference synthesis (DCS) uses image generation or editing to make the required changes while preserving asset identity and already correct attributes.
Asset review checks target attributes, identity preservation, and applicability, and validated visual references are used to generate the next candidate:
\begin{equation}
  \label{eq:repairloop}
  \begin{aligned}
    R_i^{(t+1)} &= \operatorname{Validate}\!\left(
      \operatorname{Prepare}(R_i^{(t)},N_i^{(t)} \mid O_i)\right),\\
    V_i^{(t+1)} &= F\!\left(
      C_i \mid R_i^{(t+1)},D_i^{(t)},P_i^{(t)},O_i,V_i^{(t)}\right).
  \end{aligned}
\end{equation}

Here, $R_i^{(t)}$ contains the current references, $C_i$ specifies shot constraints, $O_i$ contains accepted source or preceding evidence, $V_i^{(t)}$ is the candidate under review, and $F$ denotes the framework-level repair procedure.
The agent uses the current candidate, defects, content to preserve, and preceding evidence to revise the prompt and organize reference conditions.
The video model receives the revised prompt and validated visual references to generate the next candidate.
The generated candidate is reviewed again to check whether the intended actions and state transitions are realized.
Once the ordinary repair allowance is exhausted, ChiefDecision returns $\mathrm{extra}$ for a chief-authorized repair or $\mathrm{select}$ for compromise selection.
Each additional repair must construct, inspect, and actually use a newly generated reference targeting the current defects.
In \cref{alg:production}, Record stores the selected result's actual observations, unconfirmed attributes, and residual defects separately in set memory and retains the original reviews.
ChiefConfirm verifies evidence and applicable scope for each asset attribute. Publish updates global memory and handoff evidence with the confirmed shareable observations.
Their scope filtering and cross-set handoff mechanisms are described in \cref{sec:agents}.

\begin{table*}[t]
  \centering
  \caption{\textbf{Comprehensive long-video comparison across four benchmarks.} ST-Bench, ViMax-Bench, and EntityBench use predefined storyboards, while DirectorBench uses autonomous planning. Overall is the pre-rounding macro-average of all metrics. Higher is better. Best and second-best results are bolded and underlined.}
  \label{tab:main}
  \small
  \setlength{\tabcolsep}{2pt}
  \renewcommand{\arraystretch}{1.2}
  \setlength{\arrayrulewidth}{0.4pt}
  \setlength{\doublerulesep}{0pt}
  \setlength{\dashlinedash}{2pt}
  \setlength{\dashlinegap}{1.5pt}
  \begin{tabularx}{\textwidth}{@{}>{\centering\arraybackslash}p{10pt}>{\raggedright\arraybackslash}p{.21\textwidth}*{5}{Y}>{\centering\arraybackslash}p{35pt}}
\hline

\hline

\hline
& \multirow{2}{*}{\textbf{Method}} &
\multicolumn{1}{c}{\textbf{Visual Quality}} &
\multicolumn{2}{c}{\textbf{Text Alignment}} &
\multicolumn{2}{c}{\textbf{Cross-shot Consistency}} &
\multirow{2}{*}{\textbf{Overall}} \\
\cmidrule(lr){3-3}\cmidrule(lr){4-5}\cmidrule(lr){6-7}
& & AQ & Global & Shot & Top-10 & All & \\
\hline
& \methodlogo{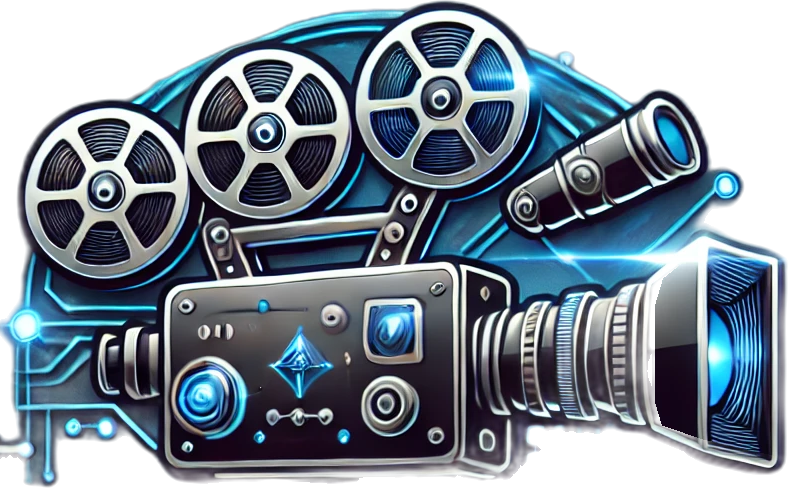}MovieAgent~\cite{wu2025movieagent} & 0.632 & \textbf{0.264} & \textbf{0.214} & 0.589 & 0.548 & 0.449 \\
& \methodlogo{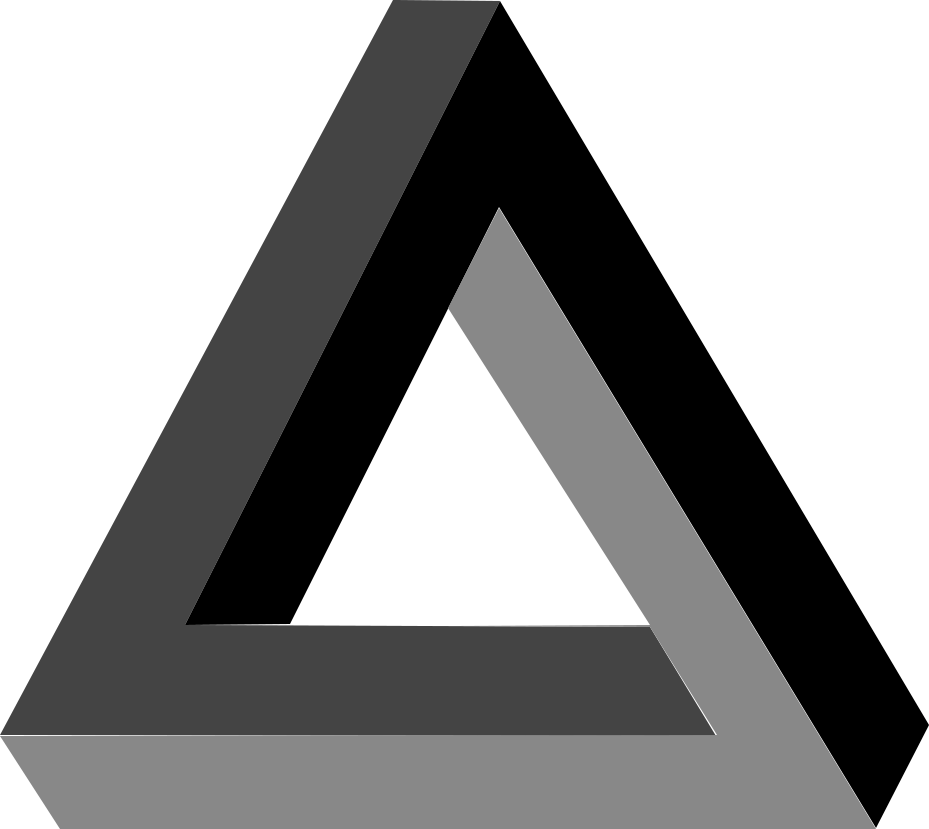}ViMax~\cite{huang2026vimax} & 0.630 & 0.254 & \underline{0.204} & 0.567 & 0.524 & 0.436 \\
& \methodlogo{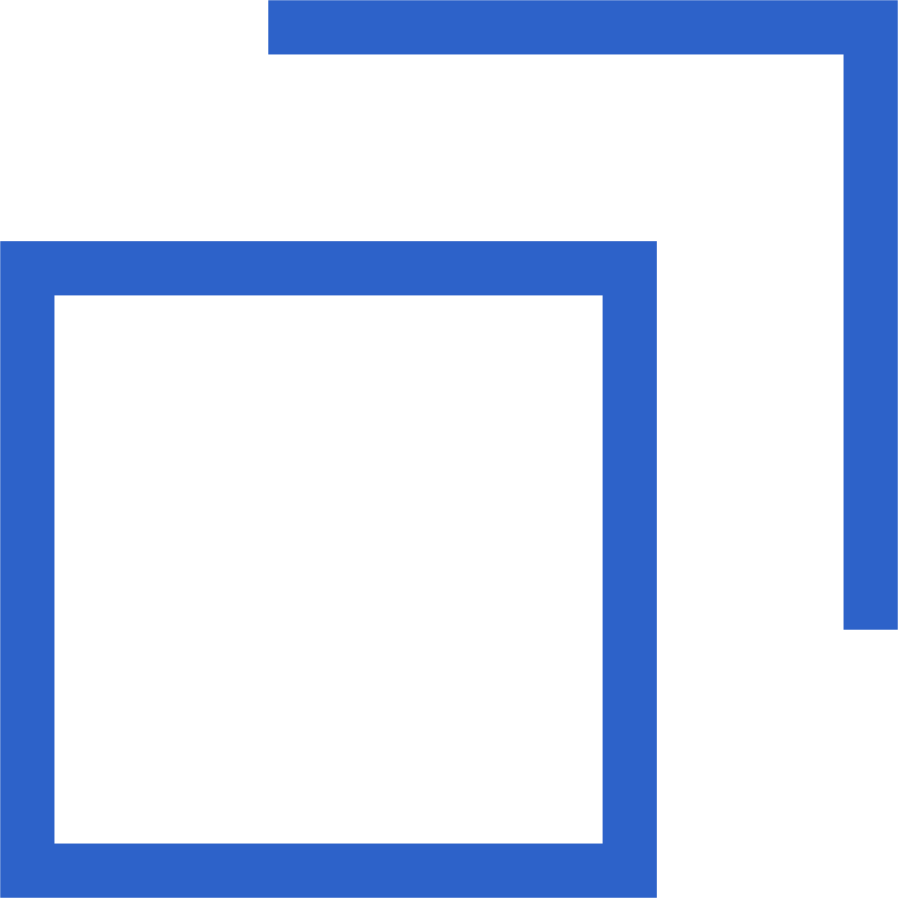}StoryEngine~\cite{wang2026storyengine} & \underline{0.642} & \underline{0.257} & 0.202 & 0.579 & 0.546 & 0.445 \\
& \methodlogo{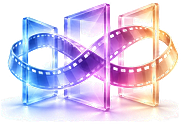}ATAS~\cite{wang2026atas} & 0.636 & 0.255 & 0.202 & \textbf{0.622} & \textbf{0.571} & \textbf{0.457} \\
\cdashline{2-8}
\rowcolor{StateWeaverRow}
\cellcolor{white}\multirow{-5}{*}{\rotatebox[origin=c]{90}{\textbf{ST-Bench}}}
& \methodlogo{stateweaver}\textbf{\method{} (Ours)} & \textbf{0.648} & 0.251 & \underline{0.204} & \underline{0.596} & \underline{0.558} & \underline{0.451} \\
\hline
\end{tabularx}\par\nointerlineskip
  \begin{tabularx}{\textwidth}{@{}>{\centering\arraybackslash}p{10pt}>{\raggedright\arraybackslash}p{.21\textwidth}*{9}{Y}>{\centering\arraybackslash}p{35pt}}
& \multirow{2}{*}{\textbf{Method}} &
\multicolumn{3}{c}{\textbf{Narrative Alignment}} &
\multicolumn{3}{c}{\textbf{Visual \& Motion Quality}} &
\multicolumn{3}{c}{\textbf{Long-range Consistency}} &
\multirow{2}{*}{\textbf{Overall}} \\
\cmidrule(lr){3-5}\cmidrule(lr){6-8}\cmidrule(lr){9-11}
& & TVA & Req. & Order & AQ & IQ & Temp. & Char. & Env. & State & \\
\hline
& \methodlogo{movieagent}MovieAgent~\cite{wu2025movieagent} & \textbf{0.576} & 0.463 & 0.675 & 0.588 & 0.737 & 0.986 & 0.412 & 0.662 & 0.475 & 0.619 \\
& \methodlogo{vimax}ViMax~\cite{huang2026vimax} & 0.574 & \textbf{0.512} & \textbf{0.750} & \underline{0.590} & \underline{0.750} & 0.989 & \underline{0.563} & \textbf{0.875} & \textbf{0.563} & \underline{0.685} \\
& \methodlogo{storyengine}StoryEngine~\cite{wang2026storyengine} & 0.574 & 0.487 & \underline{0.724} & 0.586 & \textbf{0.765} & \textbf{0.993} & 0.474 & 0.763 & 0.539 & 0.656 \\
& \methodlogo{atas}ATAS~\cite{wang2026atas} & \textbf{0.576} & 0.487 & 0.700 & 0.570 & 0.749 & \underline{0.992} & \underline{0.563} & \underline{0.863} & 0.525 & 0.669 \\
\cdashline{2-12}
\rowcolor{StateWeaverRow}
\cellcolor{white}\multirow{-5}{*}{\rotatebox[origin=c]{90}{\textbf{ViMax-Bench}}}
& \methodlogo{stateweaver}\textbf{\method{} (Ours)} & \underline{0.575} & \underline{0.500} & \textbf{0.750} & \textbf{0.598} & 0.736 & 0.991 & \textbf{0.600} & \textbf{0.875} & \underline{0.550} & \textbf{0.686} \\
\hline
\end{tabularx}\par\nointerlineskip
  \begin{tabularx}{\textwidth}{@{}>{\centering\arraybackslash}p{10pt}>{\raggedright\arraybackslash}p{.21\textwidth}*{10}{Y}>{\centering\arraybackslash}p{35pt}}
& \multirow{2}{*}{\textbf{Method}} &
\multicolumn{2}{c}{\textbf{Visual Quality}} &
\multicolumn{4}{c}{\textbf{Intra-shot Fidelity}} &
\multicolumn{4}{c}{\textbf{Cross-shot Consistency}} &
\multirow{2}{*}{\textbf{Overall}} \\
\cmidrule(lr){3-4}\cmidrule(lr){5-8}\cmidrule(lr){9-12}
& & AQ & IQ & Char. & Obj. & Scene & Action & Char. & Obj. & Scene & Bdry. & \\
\hline
& \methodlogo{movieagent}MovieAgent~\cite{wu2025movieagent} & 0.609 & 0.723 & 0.942 & 0.880 & 0.841 & 0.852 & 0.871 & 0.814 & 0.748 & 0.465 & 0.774 \\
& \methodlogo{vimax}ViMax~\cite{huang2026vimax} & \textbf{0.624} & 0.723 & \textbf{0.975} & \underline{0.923} & 0.895 & \underline{0.917} & \underline{0.886} & 0.825 & 0.775 & 0.522 & 0.806 \\
& \methodlogo{storyengine}StoryEngine~\cite{wang2026storyengine} & 0.613 & \underline{0.731} & 0.953 & 0.903 & 0.888 & 0.837 & 0.832 & 0.840 & 0.843 & 0.458 & 0.790 \\
& \methodlogo{atas}ATAS~\cite{wang2026atas} & \underline{0.620} & 0.706 & \underline{0.969} & 0.908 & \underline{0.914} & 0.874 & 0.885 & \underline{0.844} & \underline{0.898} & \underline{0.537} & \underline{0.815} \\
\cdashline{2-13}
\rowcolor{StateWeaverRow}
\cellcolor{white}\multirow{-5}{*}{\rotatebox[origin=c]{90}{\textbf{EntityBench}}}
& \methodlogo{stateweaver}\textbf{\method{} (Ours)} & 0.604 & \textbf{0.754} & \textbf{0.975} & \textbf{0.957} & \textbf{0.961} & \textbf{0.925} & \textbf{0.888} & \textbf{0.888} & \textbf{0.935} & \textbf{0.616} & \textbf{0.850} \\
\hline
\end{tabularx}\par\nointerlineskip
  \begin{tabularx}{\textwidth}{@{}>{\centering\arraybackslash}p{10pt}>{\raggedright\arraybackslash}p{.21\textwidth}*{10}{Y}>{\centering\arraybackslash}p{35pt}}
& \multirow{2}{*}{\textbf{Method}} &
\multicolumn{3}{c}{\textbf{Story Planning}} &
\multicolumn{3}{c}{\textbf{Video Fidelity}} &
\multicolumn{4}{c}{\textbf{Long-horizon Quality}} &
\multirow{2}{*}{\textbf{Overall}} \\
\cmidrule(lr){3-5}\cmidrule(lr){6-8}\cmidrule(lr){9-12}
& & Reas. & Nov. & Req. & Script & Dem. & Text & Temp. & Light. & Trans. & Stab. & \\
\hline
& \methodlogo{movieagent}MovieAgent~\cite{wu2025movieagent} & 0.631 & 0.371 & 0.910 & 0.600 & 0.761 & 0.473 & 0.562 & 0.634 & \underline{0.214} & 0.672 & 0.583 \\
& \methodlogo{vimax}ViMax~\cite{huang2026vimax} & 0.816 & \underline{0.461} & 0.904 & 0.707 & 0.778 & 0.509 & 0.459 & 0.672 & 0.142 & 0.558 & 0.601 \\
& \methodlogo{storyengine}StoryEngine~\cite{wang2026storyengine} & \underline{0.830} & 0.401 & \textbf{0.964} & 0.759 & \textbf{0.835} & 0.524 & 0.551 & \underline{0.721} & 0.211 & 0.620 & 0.642 \\
& \methodlogo{atas}ATAS~\cite{wang2026atas} & 0.797 & 0.450 & 0.944 & \underline{0.788} & 0.810 & \underline{0.550} & \underline{0.582} & 0.720 & 0.149 & \underline{0.690} & \underline{0.648} \\
\cdashline{2-13}
\rowcolor{StateWeaverRow}
\cellcolor{white}\multirow{-5}{*}{\rotatebox[origin=c]{90}{\textbf{DirectorBench}}}
& \methodlogo{stateweaver}\textbf{\method{} (Ours)} & \textbf{0.874} & \textbf{0.512} & \underline{0.957} & \textbf{0.806} & \underline{0.821} & \textbf{0.602} & \textbf{0.725} & \textbf{0.759} & \textbf{0.450} & \textbf{0.729} & \textbf{0.724} \\
\hline

\hline

\hline
\end{tabularx}
\end{table*}

\subsection{Hierarchical Collaboration}
\label{sec:agents}

\myPara{Hierarchical Contexts.}
\method defines local production scopes by sets, organizes temporary execution contexts by specialized task, clip, and repair round, and recovers production information from persistent memory.
The chief director oversees planning, task assignment, and cross-set scheduling, delegating asset preparation, repair diagnosis, candidate reassessment, and blocked-task handling to executive directors.
Task-context isolation (TCI) creates an independent context for each specialized task to read task requirements and relevant evidence, return conclusions and evidence pointers, and release the context upon completion.
Set directors apply clip- and repair-round planning-context isolation (CCI), starting a new planning context when switching clips or beginning a new repair round.
Each context reads the current plan, applicable references, confirmed observations, and relevant repair decisions. Complete conversations from the preceding task or round are not automatically carried over.
Formal self-review uses an independent candidate-review context~\cite{xu2026dontreadlog}, with review results read by the next planning round.
The chief director reads specialized reports and unresolved issues on demand, persistently recording scheduling decisions and resumption points for later coordination.

\myPara{Hierarchical Story and Asset Memory.}
\method uses global memory $M_G$ for story facts, asset information, and confirmed observations shared across sets, and set memory $M_S$ for local plans, generated observations, and repair decisions.
Both levels comprise story memory and asset memory, respectively organizing character relationships, event order, and goal changes, and visual materials and their provenance indexed by asset, story state, and reference version.
Entries record applicable events, clips, and story intervals, distinguishing planned facts, target references, and actual observations.
The system first restricts entries to accessible records with valid sources, then applies memory scope filtering (SF) by event and story interval. Agents read task-relevant plans, references, and preceding evidence from these results and expand detailed records on demand.
TCI and CCI determine where execution histories are isolated, while SF determines which entries are read from persistent memory.
Complete production records persist when temporary contexts are released.
Set directors maintain local memory, while the chief director confirms and publishes shared facts and observations with their applicable story intervals, asset scopes, and source versions before those records are released across sets.

\myPara{Dependency-Driven Parallel Generation.}
\method constructs a directed dependency graph from the states and actual visual evidence required by each task.
Tasks that inherit upstream appearance or the final state of an action declare source events, asset states, required attributes, and applicable scopes.
Tasks sharing fixed story settings can prepare independently using published references.
The system checks dependency requirements against currently valid evidence to determine the ready-task set $\mathcal Q$:
\begin{equation}
    \mathcal Q\!=\!\bigl\{T_{k,i}\!\in\!\mathcal T_{\mathrm{pending}}\!\bigm|\!\operatorname{Satisfied}(\operatorname{Dep}(T_{k,i}),\!\mathcal E_{\mathrm{acc}})\bigr\}.
  \label{eq:readiness}
\end{equation}

\begin{table*}[t]
  \centering
  \caption{\textbf{Ablation of key components in Active States and Hierarchical Collaboration.}
  RE denotes reference expansion through additions or revisions. DCS denotes defect-conditioned synthesis of reference images.
  TCI denotes task-context isolation.
  CCI denotes clip- and repair-round planning-context isolation.
  $\checkmark$ and $\times$ indicate enabled and disabled components, respectively. -- denotes not applicable.
  Category scores are means. Overall averages all DirectorBench metrics before rounding.}
  \label{tab:ablation}
  \small
  \setlength{\tabcolsep}{2pt}
  \renewcommand{\arraystretch}{1.25}
  \setlength{\arrayrulewidth}{0.4pt}
  \setlength{\doublerulesep}{0pt}
  \long\def\line#1\par{\noalign{\hrule height \arrayrulewidth}}
  \setlength{\dashlinedash}{2pt}
  \setlength{\dashlinegap}{1.5pt}
  \begin{tabularx}{\textwidth}{>{\raggedright\arraybackslash}p{0.23\textwidth}*{2}{>{\hsize=.8\hsize\linewidth=\hsize\centering\arraybackslash}X}*{2}{>{\hsize=1.35\hsize\linewidth=\hsize\centering\arraybackslash}X}*{2}{>{\hsize=.8\hsize\linewidth=\hsize\centering\arraybackslash}X}>{\hsize=1.3\hsize\linewidth=\hsize\centering\arraybackslash}X>{\hsize=.8\hsize\linewidth=\hsize\centering\arraybackslash}X}
    \line

    \line

    \line

    \multirow{2}{*}{\textbf{Method}} &
    \multicolumn{2}{c}{\textbf{Active States}} &
    \multicolumn{2}{c}{\textbf{Hierarchical Collaboration}} &
    \multicolumn{4}{c}{\textbf{Results}} \\
    \cmidrule(lr){2-3}\cmidrule(lr){4-5}\cmidrule(lr){6-9}
    & RE
    & DCS
    & TCI
    & CCI
    & \textbf{Story}
    & \textbf{Video}
    & \textbf{Long-horizon}
    & \textbf{Overall} \\
    \hline
    \method{} with Frozen Refs.
    & $\times$ & -- & $\checkmark$ & $\checkmark$ & 0.779 &	0.734  &	0.622 & 0.703 \\
    \method{} w/o DCS
    & $\checkmark$ & $\times$ & $\checkmark$ & $\checkmark$ & 0.748	& 0.727 & 	0.640 & 0.699  \\
    \hline
    \method{} w/o TCI
    & $\checkmark$ & $\checkmark$ & $\times$ & $\checkmark$ & 0.746&	0.724&	0.656 & 0.703  \\
    \method{} w/o CCI
    & $\checkmark$ & $\checkmark$ & $\checkmark$ & $\times$ & \textbf{0.793} &	0.713&	\underline{0.660} & \underline{0.716} \\
    \method{} w/o TCI \& CCI
    & $\checkmark$ & $\checkmark$ & $\times$ & $\times$ &  0.772&	0.705&	0.600 & 0.683  \\
    \hdashline
    \rowcolor{StateWeaverRow}
    \textbf{\method{} (Ours)}
    & $\checkmark$ & $\checkmark$ & $\checkmark$ & $\checkmark$ & \underline{0.781} & \textbf{0.743} & \textbf{0.666} & \textbf{0.724} \\
    \line

    \line

    \line

  \end{tabularx}
\end{table*}

Here, $T_{k,i}$ is the generation task for clip $i$ in set $k$, $\mathcal T_{\mathrm{pending}}$ is the set of pending tasks, $\operatorname{Dep}(T_{k,i})$ specifies its upstream evidence requirements, and $\mathcal E_{\mathrm{acc}}$ contains currently valid confirmed observations and handoff evidence.
$\operatorname{Satisfied}$ requires matching source events and asset states, confirmed attributes covering the current requirements, correct applicability, and valid selected source clips and reference versions.
Tasks with satisfied dependencies can execute in parallel, while unresolved tasks can begin shot planning and asset preparation that does not require upstream observations.
Once a source clip is selected, the chief director performs observed-state handoff verification (OH), checking each required downstream attribute for support from confirmed observations and verifying that story intervals, selected source clips, and reference versions match the requirements.
After verification, task inputs are bound to the observations actually used and their source versions before eligible tasks are unlocked, allowing subsequent checks to trace the evidence used by each task.
Changes to shared states, selected source clips, or reference versions trigger reassessment of affected handoffs and mark relevant unsubmitted reference plans for review.
Before replacing a reference already used in production, the chief director assesses its impact from actual usage.
Submitted requests and historical results retain their original input versions, while subsequent reassessment and repair are determined by reference dependencies.
Version binding preserves decisions and selectively reopens affected descendants without interrupting independent tasks already in progress.

\begin{table}[t]
  \centering
  \caption{\textbf{Memory filtering and state handoffs.}
  SF denotes memory scope filtering.
  OH denotes observed-state handoff verification.
  Category scores are means. Overall averages all DirectorBench metrics before rounding.}
  \label{tab:scope-filtering}
  \small
  \setlength{\tabcolsep}{1.5pt}
  \renewcommand{\arraystretch}{1.25}
  \setlength{\arrayrulewidth}{0.4pt}
  \setlength{\doublerulesep}{0pt}
  \long\def\line#1\par{\noalign{\hrule height \arrayrulewidth}}
  \setlength{\dashlinedash}{2pt}
  \setlength{\dashlinegap}{1.5pt}
  \resizebox{\linewidth}{!}{%
  \begin{tabular}{lcccc}
    \line

    \line

    \line

    \textbf{Method} & \textbf{Story} & \textbf{Video} & \textbf{Long-horizon} & \textbf{Overall} \\
    \hline
    \method{} w/o SF & \underline{0.766} & 0.727 & 0.640 & 0.704\\
    \method{} w/o OH & 0.762 & \underline{0.731}& \underline{0.662} & \underline{0.713} \\
    \method{} w/o SF \& OH & 0.742		& 0.721& 0.642 & 0.696\\
    \hdashline
    \rowcolor{StateWeaverRow}
    \textbf{\method{} (Ours)} & \textbf{0.781}	& \textbf{0.743}	& \textbf{0.666}	& \textbf{0.724} \\
    \line

    \line

    \line

  \end{tabular}%
  }
\end{table}

\section{Experiments}
\label{sec:experiments}

\subsection{Experimental Details}
\label{sec:protocol}

\myPara{Implementation details.}
StateWeaver’s screenwriter, chief director, executive directors, and set directors all use GPT-6.1-sol, with the reasoning effort set to high by default.
GPT-Image-2.5-Sunburst generates or edits visual references, and MiniMax-H3-Sol~\cite{minimax2026h3} generates videos from prompts and reference images in Ref2VA mode.
The video model runs locally on NVIDIA RTX PRO 6000 GPU hardware following the Sol-H3 deployment procedure~\cite{li2026solh3, xie2024sana}.
Detailed model and inference configurations, video specifications, and agent prompts are provided in the supplementary material, together with per-module generation and review settings.

\begin{table*}[t]
  \centering
  \caption{\textbf{GSB human evaluation.}
  G/S/B are percentages over 75 story-level comparisons per baseline and dimension. Dimension scores use pre-rounding counts, and Total Score averages the three dimensions.
  Darker orange denotes higher values.}
  \label{tab:human}
  \definecolor{GSBOrange}{HTML}{EB8F36}
  \newcommand{\gsborange}[3]{%
    \edef\gsbcellcolor{\noexpand\cellcolor{GSBOrange!\fpeval{
      round(5+33*min(1,max(0,((#1)-(#2))/max(0.0001,(#3)-(#2)))),0)}}}%
    \gsbcellcolor #1%
  }
  \small
  \setlength{\tabcolsep}{2.8pt}
  \renewcommand{\arraystretch}{1.25}
  \setlength{\arrayrulewidth}{0.4pt}
  \setlength{\doublerulesep}{0pt}
  \long\def\line#1\par{\noalign{\hrule height \arrayrulewidth}}
  \begin{tabularx}{\textwidth}{l*{3}{*{3}{>{\centering\arraybackslash}X}>{\centering\arraybackslash\bfseries}X}>{\centering\arraybackslash\bfseries}X}
    \line

    \line

    \line

    \multirow{2}{*}{\textbf{Comparison}} &
    \multicolumn{4}{c}{\textbf{Continuity}} &
    \multicolumn{4}{c}{\textbf{State Transitions}} &
    \multicolumn{4}{c}{\textbf{Visual}} &
    \multirow[c]{2}{*}[-2.5pt]{\shortstack[c]{\textbf{Total}\\\mbox{\textbf{Score}}}} \\
    \cmidrule(lr){2-5}\cmidrule(lr){6-9}\cmidrule(lr){10-13}
    & G & S & B & \makebox[\linewidth][c]{Score\,$\uparrow$}
    & G & S & B & \makebox[\linewidth][c]{Score\,$\uparrow$}
    & G & S & B & \makebox[\linewidth][c]{Score\,$\uparrow$} & \\
    \hline
    Ours vs. MovieAgent~\cite{wu2025movieagent}
    & \gsborange{76.0}{56.0}{88.0} & \gsborange{17.3}{9.3}{38.7} & \gsborange{6.7}{2.7}{6.7} & \gsborange{69.3}{50.7}{85.3} & \gsborange{56.0}{46.7}{70.7} & \gsborange{34.7}{24.0}{46.7} & \gsborange{9.3}{5.3}{9.3} & \gsborange{46.7}{37.3}{65.3} & \gsborange{41.3}{38.7}{61.3} & \gsborange{14.7}{14.7}{38.7} & \gsborange{44.0}{9.3}{44.0} & \gsborange{-2.7}{-2.7}{52.0} & \gsborange{37.8}{35.6}{67.6} \\
    Ours vs. ViMax~\cite{huang2026vimax}
    & \gsborange{56.0}{56.0}{88.0} & \gsborange{38.7}{9.3}{38.7} & \gsborange{5.3}{2.7}{6.7} & \gsborange{50.7}{50.7}{85.3} & \gsborange{46.7}{46.7}{70.7} & \gsborange{46.7}{24.0}{46.7} & \gsborange{6.7}{5.3}{9.3} & \gsborange{40.0}{37.3}{65.3} & \gsborange{38.7}{38.7}{61.3} & \gsborange{38.7}{14.7}{38.7} & \gsborange{22.7}{9.3}{44.0} & \gsborange{16.0}{-2.7}{52.0} & \gsborange{35.6}{35.6}{67.6} \\
    Ours vs. StoryEngine~\cite{wang2026storyengine}
    & \gsborange{61.3}{56.0}{88.0} & \gsborange{32.0}{9.3}{38.7} & \gsborange{6.7}{2.7}{6.7} & \gsborange{54.7}{50.7}{85.3} & \gsborange{46.7}{46.7}{70.7} & \gsborange{44.0}{24.0}{46.7} & \gsborange{9.3}{5.3}{9.3} & \gsborange{37.3}{37.3}{65.3} & \gsborange{50.7}{38.7}{61.3} & \gsborange{32.0}{14.7}{38.7} & \gsborange{17.3}{9.3}{44.0} & \gsborange{33.3}{-2.7}{52.0} & \gsborange{41.8}{35.6}{67.6} \\
    Ours vs. ATAS~\cite{wang2026atas}
    & \gsborange{88.0}{56.0}{88.0} & \gsborange{9.3}{9.3}{38.7} & \gsborange{2.7}{2.7}{6.7} & \gsborange{85.3}{50.7}{85.3} & \gsborange{70.7}{46.7}{70.7} & \gsborange{24.0}{24.0}{46.7} & \gsborange{5.3}{5.3}{9.3} & \gsborange{65.3}{37.3}{65.3} & \gsborange{61.3}{38.7}{61.3} & \gsborange{29.3}{14.7}{38.7} & \gsborange{9.3}{9.3}{44.0} & \gsborange{52.0}{-2.7}{52.0} & \gsborange{67.6}{35.6}{67.6} \\
    \line

    \line

    \line

  \end{tabularx}
\end{table*}

\myPara{Evaluation settings.}
We construct an evaluation suite of 75 text-driven video production tasks from ST-Bench~\cite{zhang2025storymem}, ViMax-Bench~\cite{huang2026vimax}, EntityBench~\cite{he2026entitybench}, and DirectorBench~\cite{chen2026directorbench}, and compare \method with MovieAgent~\cite{wu2025movieagent}, ViMax~\cite{huang2026vimax}, StoryEngine~\cite{wang2026storyengine}, and ATAS~\cite{wang2026atas}.
For the first three benchmarks, we retain the original shot content and order.
On DirectorBench, each framework plans its own shots.
All five frameworks use identical text, image generation, and video generation models, while running their respective production workflows with the same story inputs and output specifications.
Evaluation uses videos without audio and covers aesthetic quality, consistency within and across scenes, entity and action fidelity, narrative plausibility, requirement adherence, and generation stability~\cite{huang2024vbench}.
The supplementary material details the task composition of each benchmark, metric computation procedures, and the full protocol for aggregating results.

\subsection{Main Results}
\label{sec:main-results}

\myPara{Predefined storyboards.}
We compare \method with the baselines on benchmarks with predefined storyboards in \cref{tab:main}.
ST-Bench, ViMax-Bench, and EntityBench provide per-shot descriptions and shot order. \method ranks first overall on ViMax-Bench and EntityBench. On ST-Bench, it ranks second with the best aesthetics and competitive cross-shot consistency.
Multiple scene and entity consistency metrics indicate that \method maintains visual entities and their states throughout continuous production.
For video quality, \method accurately depicts the predefined shot content while maintaining good visual quality and text alignment.
These results indicate that \method can reliably extend the high-quality clip generation capabilities of foundation models to coherent long-video production while preserving competitive per-shot fidelity.

\myPara{Autonomously planned storyboards.}
\cref{tab:main} also presents the comparison with autonomously planned storyboards.
DirectorBench allows each framework to plan shots from the story requirements and complete long-video production, and \method continues to show a clear advantage.
Temporal coherence and text--video consistency metrics indicate that \method maintains cross-shot coherence and narrative alignment under autonomous planning.
Without predefined shot content or order, this setting places greater demands on the system to maintain event progression and state continuity during planning and execution, providing a stronger test of its ability to preserve continuity autonomously.
These gains in story planning, video quality, and adherence to user requirements demonstrate the advantage of \method in autonomous long-video production.

\subsection{Ablation Analysis}
\label{sec:ablations}

We conduct ablation experiments on DirectorBench to evaluate the role of each key component of \method.
Each ablation variant removes or disables only the components under investigation, while all other modules and experimental settings remain identical to those of the full \method framework, ensuring a fair comparison.

\myPara{Active States.}
To assess Active States, we ablate reference expansion (RE) and defect-conditioned reference synthesis (DCS) in \cref{tab:ablation}.
The full model scores 0.724 Overall, compared with 0.703 for frozen references and 0.699 without DCS.
The corresponding changes in Video and Long-horizon scores support the joint contribution of adaptive reference expansion and defect-conditioned synthesis, without attributing any single metric change to an individual repair behavior.
Frozen references preserve Story at 0.779 but reduce Long-horizon to 0.622, while removing DCS produces broader declines across all three categories. Together, they highlight the distinct roles of adaptive expansion and repair-specific synthesis.

\myPara{Hierarchical Collaboration.}
We evaluate task-context isolation (TCI) and clip- and repair-round planning-context isolation (CCI) in \cref{tab:ablation}.
Removing TCI or CCI lowers Overall to 0.703 and 0.716, respectively.
Their category profiles differ: removing TCI reduces Story from 0.781 to 0.746, while removing CCI reduces Video from 0.743 to 0.713.
Removing both lowers Overall to 0.683, supporting complementary roles for task-level and clip- or repair-round-level isolation.

\myPara{Memory filtering and state handoffs.}
We evaluate memory scope filtering (SF) and observed-state handoff verification (OH) in \cref{tab:scope-filtering}.
The full model scores 0.724 Overall, compared with 0.704 without SF, 0.713 without OH, and 0.696 without both.
The consistent reductions, including in Long-horizon scores, support filtering memory for the current production context and grounding handoffs in verified generation evidence.
Removing SF causes the larger Long-horizon reduction, while removing OH has the smaller Overall decline, and their joint score of 0.696 shows that the two effects are not interchangeable. The pattern is consistent with SF filtering task-relevant context and OH grounding cross-set handoffs in verified evidence.

\subsection{Human Evaluation}
\label{sec:human-eval}

In addition to automated evaluation scores, we conduct a pairwise human evaluation of complete story-level outputs.
Twenty independent raters are organized into four groups.
Each complete video pair receives five independent, fully blinded ratings from one group under the same story brief, covering cross-shot continuity, state transitions, and visual quality.
For each story--baseline pair and dimension, an outcome is counted as G, S, or B when at least three raters select the same option. Otherwise, it is unresolved and excluded from that dimension's statistics.
Stories receive equal weight.
The supplementary material provides the evaluation instructions, judgment examples, and the numbers of resolved and unresolved stories for each dimension.

We compute each dimension's net preference score as
\begin{equation}
  \mathrm{Score}_{\mathrm{GSB}}
  = 100\,\frac{n_G-n_B}{n_G+n_S+n_B}.
  \label{eq:gsb-score}
\end{equation}
Here, $n_G$, $n_S$, and $n_B$ count story-level judgments favoring \method, indicating parity, and favoring the baseline, respectively.
The score ranges from $-100$ to $100$, with $0$ indicating balanced preference.

As shown in \cref{fig:qualitative} and \cref{tab:human}, \method achieves substantial advantages in cross-shot continuity and state transitions.
Differences in visual quality are smaller, with \method performing similarly to some baselines, particularly MovieAgent~\cite{wu2025movieagent} and ViMax~\cite{huang2026vimax}.
Both frameworks generate high-quality video keyframes using image models, helping explain the smaller visual-quality gap.

\section{Conclusion}
\label{sec:conclusion}

In this paper, we introduce \method, combining review-driven Active States with Hierarchical Collaboration for isolated planning and repair.
Binding confirmed observations to asset states, story intervals, and source versions enables evidence-based handoffs and coherent reuse.

\section*{LLM Usage Statement}
We use ChatGPT to assist with organizing related literature, language polishing and translation, LaTeX typesetting, and the preparation of figures and tables.
Our method and experiments also use GPT-6.1-sol, GPT-Image-2.5-Sunburst, and MiniMax-H3-Sol for agent decision-making, visual asset construction, and video generation, respectively, with configurations described in \cref{sec:protocol}.
The authors take full responsibility for final research decisions, content approval, and result interpretation, and for the paper's accuracy, originality, and citation reliability.

\appendix
\section*{Appendix}
\begingroup
\setcounter{dbltopnumber}{3}
\captionsetup[figure]{position=bottom,skip=3pt,belowskip=0pt}
\setlength{\dblfloatsep}{6pt plus 1pt minus 1pt}
\setlength{\floatsep}{9pt plus 2pt minus 1pt}
\makeatletter
\setlength{\@fptop}{0pt}
\setlength{\@dblfptop}{0pt}
\setlength{\dbltextfloatsep}{6pt plus 1pt minus 1pt}
\makeatother

\makeatletter
\newcommand{\sectioncaption}[1]{%
  \edef\suppfloatcounter{\@captype-#1}%
  \ifcsname c@\suppfloatcounter\endcsname\else
    \newcounter{\suppfloatcounter}%
  \fi
  \stepcounter{\suppfloatcounter}%
  \expandafter\edef\csname the\@captype\endcsname{%
    \getrefnumber{#1}.\arabic{\suppfloatcounter}}%
  \caption}
\makeatother

\makeatletter\global\@dbltopnum\z@\makeatother

\begin{figure*}[t]
  \centering
  \includegraphics[width=.95\textwidth]{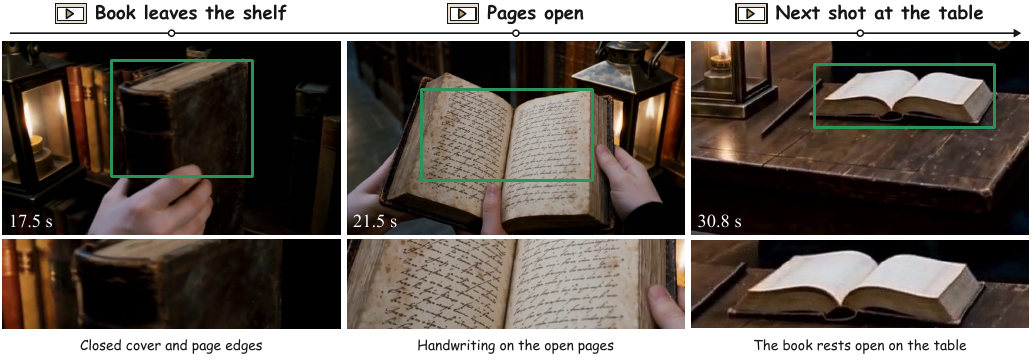}
  \sectioncaption{sec:sup-method}{\textbf{Evidence for two requirements within one event.} The details in the lower row show whether the book is open or closed. Verifying motion of the handwriting requires positional evidence across frames.}
  \label{fig:sup-book-state}
\end{figure*}

\begin{figure*}[t]
  \centering
  \includegraphics[width=.95\textwidth]{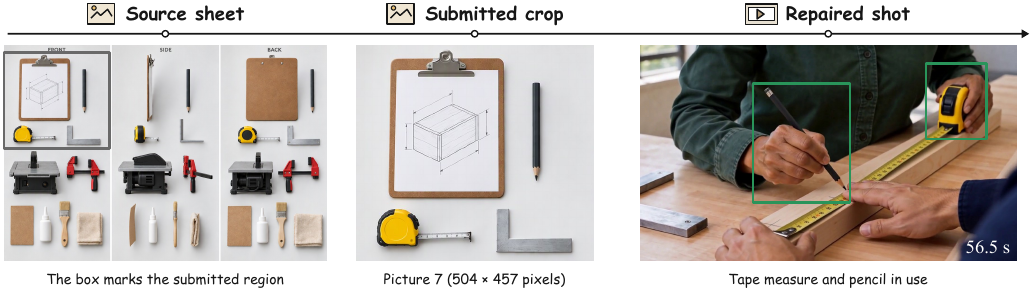}
  \sectioncaption{sec:sup-method}{\textbf{Tracing a repair input to the original reference.} The box on the left marks the crop. The center shows the image submitted as Picture 7, and the right shows the shot generated with that reference.}
  \label{fig:sup-repair-trace}
\end{figure*}

\begin{figure*}[t]
  \centering
  \includegraphics[width=.95\textwidth]{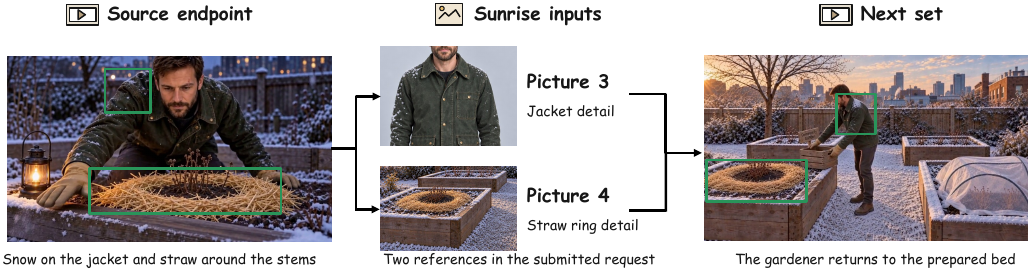}
  \sectioncaption{sec:sup-method}{\textbf{One source frame supports two downstream references.} The two center panels show details from the character and scene references submitted for the target shot. They correspond to the snow on the jacket and the straw ring, respectively.}
  \label{fig:sup-evidence-propagation}
\end{figure*}

\section{Overview}
\label{sec:sup-overview}

In this appendix, we explain how \method records event completion and visible asset states separately, traces the reference images submitted for a repair, and links specific observations in a source frame to the reference revisions used by the next set.
For reproducibility, we detail deployment and the automated and blind human evaluations on all four benchmarks under their respective storyboard settings.
We then present more representative examples and several failure cases.
Finally, we briefly discuss current limitations and explore how \method could support filmmaking, film education, and the creation of interaction video datasets for embodied AI.

\section{Additional Method Details}
\label{sec:sup-method}

\subsection{Event Completion and Visible Asset States}
\label{sec:sup-adjudication}

A storyboard event can specify both the state an object should be in at the end of a shot and an action that should occur during it.
During review, \method records the start and end times of the event in the candidate clip and the frames that support the judgment.
Alongside the completion status of the event as a whole, the review records the initial states and subsequent changes visible in the footage for each requirement.
The evidence needed depends on the attribute.
In~\cref{fig:sup-book-state}, a single frame establishes that the book is open, but verifying independent motion of the handwriting requires comparing the positions of the strokes against fixed marks on the paper across frames.
The review confirms the open book, while handwriting motion and the full event remain unconfirmed.

\begin{table}[!t]
\begin{minipage}{\columnwidth}
  \captionsetup{type=table}
\centering
  \small
  \setlength{\tabcolsep}{3pt}
  \renewcommand{\arraystretch}{1.12}
  \sectioncaption{sec:sup-implementation}{\textbf{Model configuration and runtime environment of \method for the formal evaluation batch.}}
  \label{tab:sup-models}
  \begin{tabularx}{\columnwidth}{@{}p{0.44\columnwidth}>{\raggedright\arraybackslash}X@{}}
    \hline

    \hline

    \hline
    \textbf{Component} & \textbf{Configuration} \\
    \hline
    Text and visual reasoning & GPT-6.1-sol (high) \\
    Image generation/editing & GPT-Image-2.5-Sunburst \\
    Image quality & High \\
    Image resolution & $1536\times1024$ \\
    Video generation & MiniMax-H3-Sol (Ref2VA) \\
    Output video & $1344\times768$, 24 fps \\
    Video inference GPU & NVIDIA RTX PRO 6000 \\
    Python & 3.12.8 \\
    FastAPI & 0.141.1 \\
    SQLite & 3.48.0 \\
    \hline

    \hline

    \hline
  \end{tabularx}
\end{minipage}
\par\medskip
\begin{minipage}{\columnwidth}
  \captionsetup{type=table}
\centering
  \small
  \setlength{\tabcolsep}{3pt}
  \renewcommand{\arraystretch}{1.12}
  \sectioncaption{sec:sup-implementation}{\textbf{Candidate generation, repair, and memory retrieval settings in \method.}}
  \label{tab:sup-agent-config}
  \begin{tabularx}{\columnwidth}{@{}>{\raggedright\arraybackslash}p{0.55\columnwidth}>{\raggedright\arraybackslash}X@{}}
    \hline

    \hline

    \hline
    \textbf{Setting} & \textbf{Value} \\
    \hline
    Context page limit & 8,000 UTF-8 bytes \\
    Initial candidates & 1/clip \\
    Standard repairs & 1/clip, 2/reference \\
    Extra repairs (chief director) & 2/clip, 2/reference \\
    Regular sampling & 1 fps \\
    Boundary sampling & 4 fps \\
    Frames per grid & 4 \\
    Images per call & $\leq 6$ \\
    \hline

    \hline

    \hline
  \end{tabularx}
\end{minipage}
\par\medskip
\begin{minipage}{\columnwidth}
  \captionsetup{type=table}
\centering
  \small
  \setlength{\tabcolsep}{3pt}
  \renewcommand{\arraystretch}{1.12}
  \sectioncaption{sec:sup-protocol}{\textbf{Story counts and storyboard settings in the formal evaluation batch.}}
  \label{tab:sup-benchmark-composition}
  \begin{tabularx}{\columnwidth}{@{}l>{\centering\arraybackslash}Xr@{}}
    \hline

    \hline

    \hline
    \textbf{Benchmark} & \textbf{Storyboard setting} & \textbf{Tasks} \\
    \hline
    ST-Bench~\cite{zhang2025storymem} & Predefined & 20 \\
    ViMax-Bench~\cite{huang2026vimax} & Predefined & 20 \\
    EntityBench~\cite{he2026entitybench} & Predefined & 10 \\
    DirectorBench~\cite{chen2026directorbench} & Autonomous & 25 \\
    \hline
    Total & & 75 \\
    \hline

    \hline

    \hline
  \end{tabularx}
\end{minipage}
\end{table}

\subsection{Repair Reference Crops and Input Numbering}
\label{sec:sup-reference-repair}

A repair prompt refers to its input images by number.
To trace each number to the image the generator received, \method records the source revision and selected view for each reference.
When only part of the original image is submitted, the record also includes the pixel coordinates of the crop in the original image and a content hash of the cropped image.
Images are numbered in submission order for each request.
In the request illustrated in~\cref{fig:sup-repair-trace}, the final frame of the preceding shot occupies the first position, so the sixth asset reference appears as Picture~7 in the prompt.
The input number and crop coordinates identify the precise region of the original tool sheet used for this repair.

\subsection{Observations to References Across Sets}
\label{sec:sup-evidence}

When a single source frame provides evidence about both a character and a scene, \method records a separate observation for each asset.
Each record identifies the selected candidate clip and the time of the frame within it.
It also specifies the supported attributes, the corresponding asset state, and the downstream shots and story interval to which the observation applies.
Once the chief director verifies an attribute and its scope, the receiving set links any reference inheriting that attribute to the observation.
In~\cref{fig:sup-evidence-propagation}, the character reference (Picture~3) is linked to snow on the shoulders and sleeves of the jacket, and the scene reference (Picture~4) to the straw ring in the garden bed.
Tracing either reference back identifies the same source frame and the particular detail it supports.

\section{Implementation Details}
\label{sec:sup-implementation}

\subsection{Models and Infrastructure}
\label{sec:sup-models}

We report the models, output specifications, and runtime environment used by \method in~\cref{tab:sup-models}.

\subsection{Agent Configuration}
\label{sec:sup-agent-config}

We list review settings and repair limits in~\cref{tab:sup-agent-config}. The chief director must authorize each extra repair, which uses a new, validated reference targeting the defect.

\subsection{Prompts}
\label{sec:sup-prompts}

We provide the seven system prompts used across agent roles and execution stages in~\cref{fig:sup-role-prompts-1,fig:sup-role-prompts-2,fig:sup-role-prompts-3}.

\section{Experimental Protocol}
\label{sec:sup-protocol}

\subsection{Benchmark Composition and Task Selection}
\label{sec:sup-benchmarks}

\begin{table*}[!t]
  \centering
  \small
  \setlength{\tabcolsep}{5pt}
  \renewcommand{\arraystretch}{1.12}
  \sectioncaption{sec:sup-protocol}{\textbf{Identifiers of the 75 selected production tasks.}
  ViMax-Bench entries are official story IDs. 
  Each EntityBench entry is a unique substring of an original filename. In \texttt{\detokenize{run34__i0_j23}}, \texttt{run34} identifies the source sequence, and 0 and 23 are the window boundary indices.}
  \label{tab:sup-original-ids}
  \newlength{\supTaskGroupWidth}
  \settowidth{\supTaskGroupWidth}{Multi-character interaction (C)}
  \renewcommand{\tabularxcolumn}[1]{m{#1}}
  \newcommand{\supTaskRows}[1]{%
    \begin{tabular}[c]{@{}>{\raggedright\arraybackslash}m{\supTaskGroupWidth}%
      >{\raggedright\arraybackslash}m{\dimexpr\linewidth-\supTaskGroupWidth-2\tabcolsep\relax}@{}}%
      #1%
    \end{tabular}}
  \begin{tabularx}{\textwidth}{@{}lX@{}}
    \hline

    \hline

    \hline
    \textbf{Benchmark} & \supTaskRows{%
      \textbf{Group or ID format} & \textbf{Selected tasks}} \\
    \hline
    ST-Bench~\cite{zhang2025storymem} & \supTaskRows{%
      Script name & \texttt{\detokenize{artist}}, \texttt{\detokenize{astronaut}}, \texttt{\detokenize{bean}}, \texttt{\detokenize{black}}, \texttt{\detokenize{cat}}, \texttt{\detokenize{dog}}, \texttt{\detokenize{explorer}}, \texttt{\detokenize{harry}}, \texttt{\detokenize{jane}}, \texttt{\detokenize{japan}}, \texttt{\detokenize{jobs}}, \texttt{\detokenize{little_sheldon}}, \texttt{\detokenize{neo}}, \texttt{\detokenize{phd}}, \texttt{\detokenize{qin}}, \texttt{\detokenize{queen}}, \texttt{\detokenize{robinson}}, \texttt{\detokenize{robot}}, \texttt{\detokenize{superman}}, \texttt{\detokenize{titanic}}} \\
    \hline
    ViMax-Bench~\cite{huang2026vimax} & \supTaskRows{%
      Character persistence (A) & \texttt{2}, \texttt{3}, \texttt{4}, \texttt{8}, \texttt{9}, \texttt{10} \\
      Background persistence (B) & \texttt{15}, \texttt{16}, \texttt{17}, \texttt{18}, \texttt{22}, \texttt{24} \\
      Multi-character interaction (C) & \texttt{26}, \texttt{27}, \texttt{28}, \texttt{29}, \texttt{30}, \texttt{32}, \texttt{33}, \texttt{35}} \\
    \hline
    EntityBench~\cite{he2026entitybench} & \supTaskRows{%
      Medium & \texttt{\detokenize{run34__i0_j23}}, \texttt{\detokenize{run152__i5_j56}}, \texttt{\detokenize{run317__i25_j60}}, \texttt{\detokenize{run338__i25_j34}}, \texttt{\detokenize{run367__i60_j68}}, \texttt{\detokenize{run406__i40_j66}} \\
      Hard & \texttt{\detokenize{run238__i0_j34}}, \texttt{\detokenize{run247__i0_j47}}, \texttt{\detokenize{run264__i50_j98}}, \texttt{\detokenize{run298__i40_j64}}} \\
    \hline
    DirectorBench~\cite{chen2026directorbench} & \supTaskRows{%
      Prefix \texttt{DB\_} & \texttt{\detokenize{001}}, \texttt{\detokenize{005}}, \texttt{\detokenize{012}}, \texttt{\detokenize{014}}, \texttt{\detokenize{016}}, \texttt{\detokenize{017}}, \texttt{\detokenize{018}}, \texttt{\detokenize{022}}, \texttt{\detokenize{024}}, \texttt{\detokenize{027}}, \texttt{\detokenize{031}}, \texttt{\detokenize{036}}, \texttt{\detokenize{039}},\newline \texttt{\detokenize{040}}, \texttt{\detokenize{042}}, \texttt{\detokenize{043}}, \texttt{\detokenize{044}}, \texttt{\detokenize{047}}, \texttt{\detokenize{058}}, \texttt{\detokenize{061}}, \texttt{\detokenize{063}}, \texttt{\detokenize{064}}, \texttt{\detokenize{066}}, \texttt{\detokenize{071}}, \texttt{\detokenize{072}}} \\
    \hline

    \hline

    \hline
  \end{tabularx}
\end{table*}

We selected 75 production tasks from four benchmarks, using the story as the unit of selection.
For benchmarks with predefined storyboards, we used the original scripts to select stories that vary in how shots change, how many scenes they span, and how characters interact.
For autonomous storyboarding, we first identified briefs that leave shot planning to the framework, then selected briefs covering different types of production.
All five frameworks received the same 50 tasks with predefined storyboards and 25 tasks requiring autonomous planning.
We report task counts by benchmark in~\cref{tab:sup-benchmark-composition} and list each task's original benchmark identifier in~\cref{tab:sup-original-ids}.

\myPara{ST-Bench~\cite{zhang2025storymem}.}
We selected 20 of the 30 original scripts, considering story genre, changes in viewpoint and setting, and interactions between characters and props.
Each selected story contains 7--12 shots.
We retained the original shot order and cut or continuation markers to assess whether characters and props remain recognizable when shot scale or viewpoint changes, and whether an action marked as continuous carries naturally from one shot to the next.

\myPara{ViMax-Bench~\cite{huang2026vimax}.}
We selected 20 of the 35 stories across the benchmark's three consistency categories, with six stories testing character persistence (Type~A), six testing background persistence (Type~B), and eight testing multi-character interaction (Type~C).
Each category places a different demand on cross-shot generation.
A gardener appearing across four seasons must remain recognizable as the surroundings change.
A board game tournament held in a single caf\'{e} requires the spatial layout to remain consistent across camera positions.
In a band rehearsal, the performers must remain distinguishable both when shown in alternating shots and when playing together.

\myPara{EntityBench~\cite{he2026entitybench}.}
We grouped the 140 candidate clips by source story, kept one clip per source, and selected 10 scripts based on the benchmark's difficulty labels and sequence lengths.
The six medium-difficulty scripts contain 15--22 shots each, and the four hard scripts contain 50 shots each.
We checked the shot boundaries used for continuity evaluation against the actions specified for each shot, the scheduled appearances of entities, and the cut markers.
When choosing replacements, we excluded scripts that explicitly called for depictions of graphic injury.
This range of sequence lengths lets us assess both continuity between adjacent shots and the consistency of characters, props, or locations that reappear after several intervening shots.

\myPara{DirectorBench~\cite{chen2026directorbench}.}
We reviewed all 75 original production briefs individually and identified 33 that support autonomous shot planning from text alone.
We excluded briefs that depend on external images, already specify multiple shots in detail, explicitly require an animation style, or request only a still image.
We retained individual camera-movement or visual requirements when they constrain what appears on screen while leaving the framework free to decide how to divide the sequence into shots and arrange them.
From these candidates, we selected 25 briefs based on production type and duration requirements, limiting overlap among similar tasks.
The selection covers eight categories spanning action, cinematography, commercial marketing, education and news, music, narrative, science fiction, and everyday vlogs.

\subsection{Comparative Runs and Video Inclusion}
\label{sec:sup-baselines}

For each story, we retain the original brief and visual requirements, the instructions sent to each framework, and the annotations used for scoring.
The benchmark determines the storyboard setting. All five frameworks use the same text, image, and video models.

For predefined storyboards, we evaluate the original benchmark shots.
Each framework receives the ordered shot descriptions, together with any cut or continuation markers supplied by the benchmark, and generates clips using its own workflow.
We concatenate multiple clips for the same shot in delivery order before evaluation.
The scorer matches each output to its original text by shot identifier and checks that the content agrees.

On DirectorBench, each framework plans the shot content and order from the same brief.
Each task includes the story, visual requirements, and any explicit duration constraints.
Frameworks submit the finished video and the verifiable script and storyboard used to produce it.
Using shot or clip identifiers, the scorer matches storyboard entries to measured clip durations to build the text--video alignment timeline.

\subsection{Automated Evaluation Procedures}
\label{sec:sup-metrics}

We evaluate the finished videos without audio on all four benchmarks.
For predefined storyboards, we align clips and text by original shot identifier.
For autonomous storyboards, we align the generated storyboard with the delivered video timeline.
For each method, we average each metric over stories with a valid score. Overall is the unrounded macro-average of these metric means.

\subsubsection{ST-Bench}
\label{sec:sup-metrics-st}

We sample one frame at the midpoint of each of eight equal intervals per delivered shot.
We resize the shorter side to 224 pixels and center-crop each frame before encoding it with ViCLIP (InternVid-10M-FLT)~\cite{wang2024internvid}.
We split the text into consecutive chunks of at most 30 content tokens and encode each independently.
For a story with $n$ shots, let $v_i$ denote shot $i$'s video feature, $t_{ij}$ its $j$-th prompt chunk's text feature, and $\alpha_{ij}$ that chunk's share of the prompt's content tokens.
We weight cosine similarities by token count within shots, $q_i=\sum_j\alpha_{ij}\cos(v_i,t_{ij})$, and give shots equal weight.
For global alignment, we pool video features by measured shot durations $d_i$ and compare them with the synopsis feature $t_s$, pooled by token count.
\begin{equation}
\textstyle \mathrm{Shot}=n^{-1}\sum_i q_i,\quad \mathrm{Global}=\cos\bigl(\sum_i d_i v_i,t_s\bigr).
\label{eq:sup-st-alignment}
\end{equation}

For cross-shot consistency, we form all unordered pairs of distinct shots within a story. \textit{All} uses every pair.
For \textit{Top-10}, we pool the chunk features of each shot prompt by token count and select up to ten pairs with the highest text similarity.
For either set of selected pairs $P$, we compute
\begin{equation}
C(P)=\frac{1}{|P|}\sum_{(i,j)\in P}\cos(v_i,v_j).
\label{eq:sup-st-consistency}
\end{equation}

Thus, \textit{Top-10} selects pairs using the storyboard text and scores them using the corresponding video features.
For aesthetic quality, we extract CLIP ViT-L/14~\cite{radford2021clip} features from every delivered frame, score each frame with the VBench~\cite{huang2024vbench} linear aesthetic head, and average across all frames in the story.

\subsubsection{ViMax-Bench}
\label{sec:sup-metrics-vimax}

We assess each video on narrative alignment, visual and motion quality, and long-range consistency.
Text--video alignment (TVA) maps the ST-Bench shot alignment cosine score to $[0,1]$.
The other semantic criteria cover requirement fulfillment, event order, and character, environment, and state consistency.
We uniformly sample eight frames per shot and submit them with their descriptions in storyboard order.
GPT-6.1-sol rates each criterion from 1 to 5 under a common rubric.
For five videos with no valid API result, Codex used the same scale with method identities hidden.

We sample 32 frames uniformly from the full video and score aesthetic quality with VBench and imaging quality with MUSIQ~\cite{ke2021musiq}.
For temporal smoothness, we sample a continuous 24 fps window near each of the 20\%, 40\%, 60\%, and 80\% points.
Each spans one sixteenth of the video duration, bounded between 0.3 and 0.75 seconds.
We use AMT~\cite{li2023amt} frame interpolation to score motion smoothness in each window and average the four scores when all are valid.
For shot alignment $c$, semantic rating $r$, and raw MUSIQ score $u$, we report TVA as $(c+1)/2$, the semantic score as $(r-1)/4$, and IQ as $u/100$.
For window scores $h_k$, temporal smoothness is
\begin{equation}
\mathrm{Temp}=\tfrac14\sum_{k=1}^{4}h_k.
\label{eq:sup-vimax-scales}
\end{equation}

A ViMax-Bench film receives an overall score when all nine metrics are available.
The narrative group comprises TVA, requirement fulfillment, and event order.
The visual and motion group comprises aesthetic quality (AQ), imaging quality (IQ), and temporal smoothness.
The long-range group comprises character, environment, and state consistency.
With $g_1$, $g_2$, and $g_3$ denoting the respective group means, the story-level score is
\begin{equation}
\mathrm{Overall}=(g_1+g_2+g_3)/3.
\label{eq:sup-vimax-overall}
\end{equation}

\subsubsection{EntityBench}
\label{sec:sup-metrics-entity}

We match delivered clips to the original shot annotations.
VBench supplies aesthetic and imaging quality scores.
IQ is the raw MUSIQ score divided by 100.
For intra-shot fidelity, Grounding DINO~\cite{liu2024groundingdino} locates entities matching the descriptions.
We select representative visible regions.
The evaluator scores only visible attributes using criteria specific to characters, objects, or scenes.
We divide each 1--10 rating by 10.
Within each category, we average eligible entities per shot and then weight eligible shots equally within the story.

For each delivered clip, we arrange six sampled frames chronologically in a $2\times3$ grid and submit it with valid entity boxes, using GPT-6.1-sol as the primary evaluator.
The evaluator rates overall action fidelity from the images and the original action description.
We assess object interaction only when the description explicitly involves an object.
The evaluator assigns the overall action score independently, and we divide it by 10.

Across shots, we compare only appearances annotated as the same entity.
For characters and props, we compare a representative crop with other eligible crops.
For scenes, we pair comparable views from different appearances.
We exclude a character or prop appearance if its available intra-shot fidelity score is below 0.5.
A visual evaluator judges similarity between eligible crops.
For scenes, we aggregate valid pairs within each location before averaging across locations.

We evaluate boundaries only between adjacent clips marked as continuous in the storyboard.
Let $B$ be the set of valid boundaries.
For $(i,j)\in B$, $f_{i,\mathrm{end}}$ and $f_{j,\mathrm{start}}$ are the DINOv2~\cite{oquab2024dinov2} features of the preceding clip's last frame and the next clip's first frame.
The boundary score is
\begin{equation}
\mathrm{Bdry}=\frac{1}{|B|}\sum_{(i,j)\in B}\cos(f_{i,\mathrm{end}},f_{j,\mathrm{start}}).
\label{eq:sup-entity-boundary}
\end{equation}
This distinguishes within-shot fidelity, entity consistency across appearances, and continuity at clip boundaries.
We leave scores missing when comparable evidence is unavailable and record technical failures separately.

\subsubsection{DirectorBench}
\label{sec:sup-metrics-director}

DirectorBench assesses each framework's actual script, storyboard, and finished video without audio.
We rate script plausibility, novelty, and alignment against the original brief.
We sample visual evidence within each detected shot of the finished video.
For a shot spanning $a_i$ to $b_i$, the target time for sample $k$ is
\begin{equation}
t_{ik}=a_i+\frac{k-0.5}{8}(b_i-a_i),\quad k=1,\ldots,8.
\label{eq:sup-director-sampling}
\end{equation}

We select the frame in the shot with the nearest presentation timestamp.
We take distinct frames in order when enough are available.
Short shots may reuse decoded source frames.
Using these frames, the visual evaluator, primarily GPT-6.1-sol, assesses adherence to the script and brief, temporal coherence, lighting, and generation stability.
For long videos, batches share time anchors, and the evaluator pools their observations to score the full video.

Text--video consistency combines the visual evaluator's checkpoint score with MobileViCLIP~\cite{yang2025mobileviclip} similarity.
We pair each detected shot with the text of all delivered clips overlapping its interval, in temporal order.
We encode its eight frames in order with MobileViCLIP and compare them with the matched text.
With $\bar c$ the mean shot-level cosine similarity and $V$ the evaluator's text alignment score, the result is
\begin{equation}
\mathrm{Text}=0.6V+0.4(\bar c+1)/2.
\label{eq:sup-director-text}
\end{equation}
The score requires both components.
We leave text--video consistency unscored when the timeline provides no verifiable match between the text and the video.

We assess cut quality separately using the adjacent source frames on either side of each cut.
We convert structural similarity, histogram difference, and optical flow to scores for which higher is better and weight them by 0.5, 0.25, and 0.25, respectively.
We visually review up to eight cuts scoring below 0.6.
For each confirmed discontinuity within a scene, we deduct 0.08 from the video's mean cut score and clip the result to $[0,1]$.
DirectorBench Overall weights all ten metrics equally, with three for story, three for video, and four for long-horizon quality.

\subsection{Human Evaluation Protocol}
\label{sec:sup-human-protocol}

We divide 20 independent raters into four groups.
For stories with complete videos from all five frameworks, we pair the \method video with each of the four baselines.
Five raters from the same group independently judge each pair blind.
Each reads the fixed story brief, watches both muted videos in full, and judges three dimensions separately.
As shown in~\cref{fig:sup-human-interface}, we label the two videos A and B for blind evaluation.
A keyed hash of the rater and pair identifiers determines the left--right order, which stays the same on a repeat visit.

Cross-shot continuity assesses whether characters, props, and scenes remain coherent across cuts.
State transition assesses whether key changes occur and persist in later shots.
Visual quality covers overall clarity, motion, and visible artifacts.
For example, one video may show smoother motion yet lose a prop that should persist across a cut.
A rater may prefer it for visual quality and the other for cross-shot continuity.

\begin{figure*}[t]
  \centering
  \includegraphics[width=.95\textwidth]{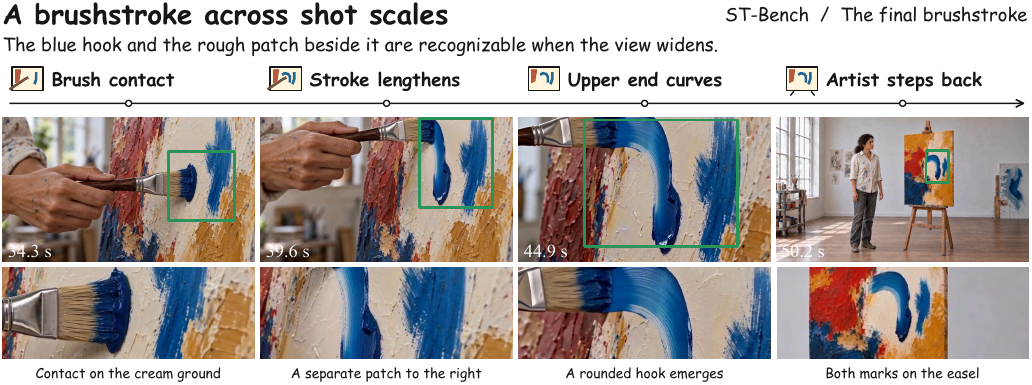}
  \sectioncaption{sec:sup-qualitative}{\textbf{Canvas appearance across shots under a predefined storyboard.}}
  \label{fig:sup-qual-predefined}
\end{figure*}

For each dimension, raters choose the left video, a tie, or the right video.
Using the order shown to each rater, we map votes to \method preferred (G), a tie (S), or baseline preferred (B).
A tally requires five valid blind votes for the same story, baseline, and video versions.
Replacing a video triggers reevaluation.
We exclude votes from raters who have seen the method identities for that story.

Once all five votes are in, at least three must agree to yield a story-level verdict for a dimension.
For example, three votes for A, one for B, and one tie give A the win.
Two votes for each video and one tie leave that dimension unresolved.
Three or more explicit ties yield S.
We weight resolved stories equally and exclude unresolved outcomes from the denominator for that dimension.
The net preference score is
\begin{equation}
\mathrm{Score}_{\mathrm{GSB}}=100\,\frac{n_G-n_B}{n_G+n_S+n_B}.
\label{eq:sup-gsb}
\end{equation}
Here $n_G$, $n_S$, and $n_B$ count the resolved stories in each category.
The total score averages the three unrounded dimension scores. For all four baselines, every pair with five valid votes reached a majority on all three dimensions, giving zero unresolved outcomes in each case. 

\section{Additional Qualitative Results}
\label{sec:sup-qualitative}

We examine cross-shot continuity in predefined and autonomously planned storyboards, focusing on whether \method preserves appearance across changes in shot scale and carries event-induced state changes into subsequent shots.
We then use the repair record for a woodworking sequence to trace which observations were verified during review and reused in a later shot.
Finally, we present additional examples of successful and failed generations.

\subsection{Predefined Storyboards}
\label{sec:sup-qual-predefined}

The ST-Bench painting storyboard calls for a cut from a close-up of the artist applying paint to a wide shot of the easel.
In the enlarged regions of~\cref{fig:sup-qual-predefined}, the flat brush leaves a new stroke with a hooked upper end to the left of an existing blue patch.
Both the hook and its position relative to the patch remain recognizable in the wide shot.
Across this change in shot scale, \method preserves both the new detail introduced by the brushstroke and the existing appearance of the canvas.
\begin{figure*}[t]
  \centering
  \includegraphics[width=.95\textwidth]{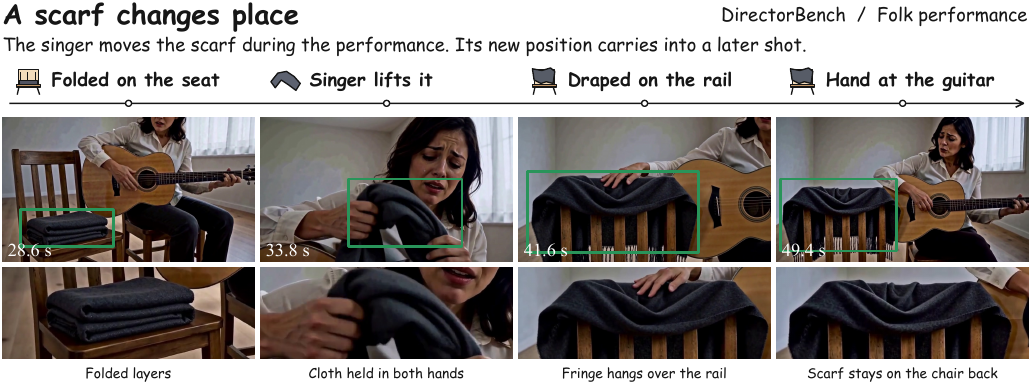}
  \sectioncaption{sec:sup-qualitative}{\textbf{Prop state transition under an autonomously planned storyboard.}}
  \label{fig:sup-qual-autonomous}
\end{figure*}

\begin{figure*}[t]
  \centering
  \includegraphics[width=.95\textwidth]{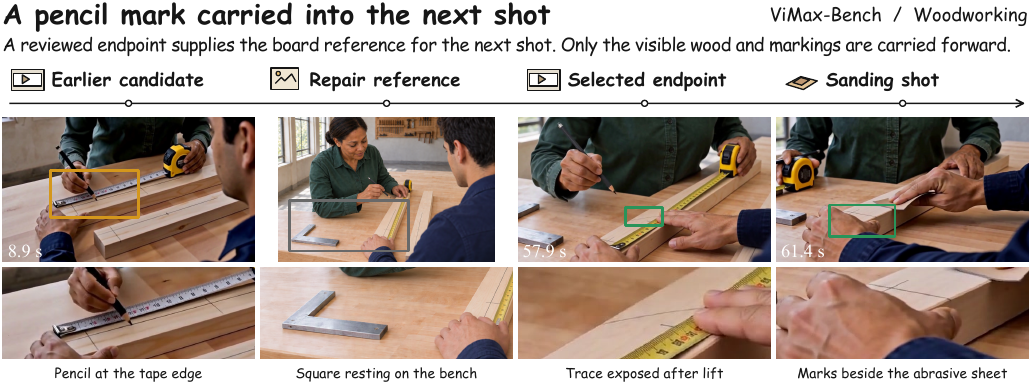}
  \sectioncaption{sec:sup-qualitative}{\textbf{Local repair and evidence reuse.} From left to right, the columns show the final frame of the early candidate, the reference used for repair, the final frame of the selected candidate, and the sanding shot.}
  \label{fig:sup-qual-repair}
\end{figure*}

\begin{figure*}[t]
  \centering
  \includegraphics[width=.95\textwidth]{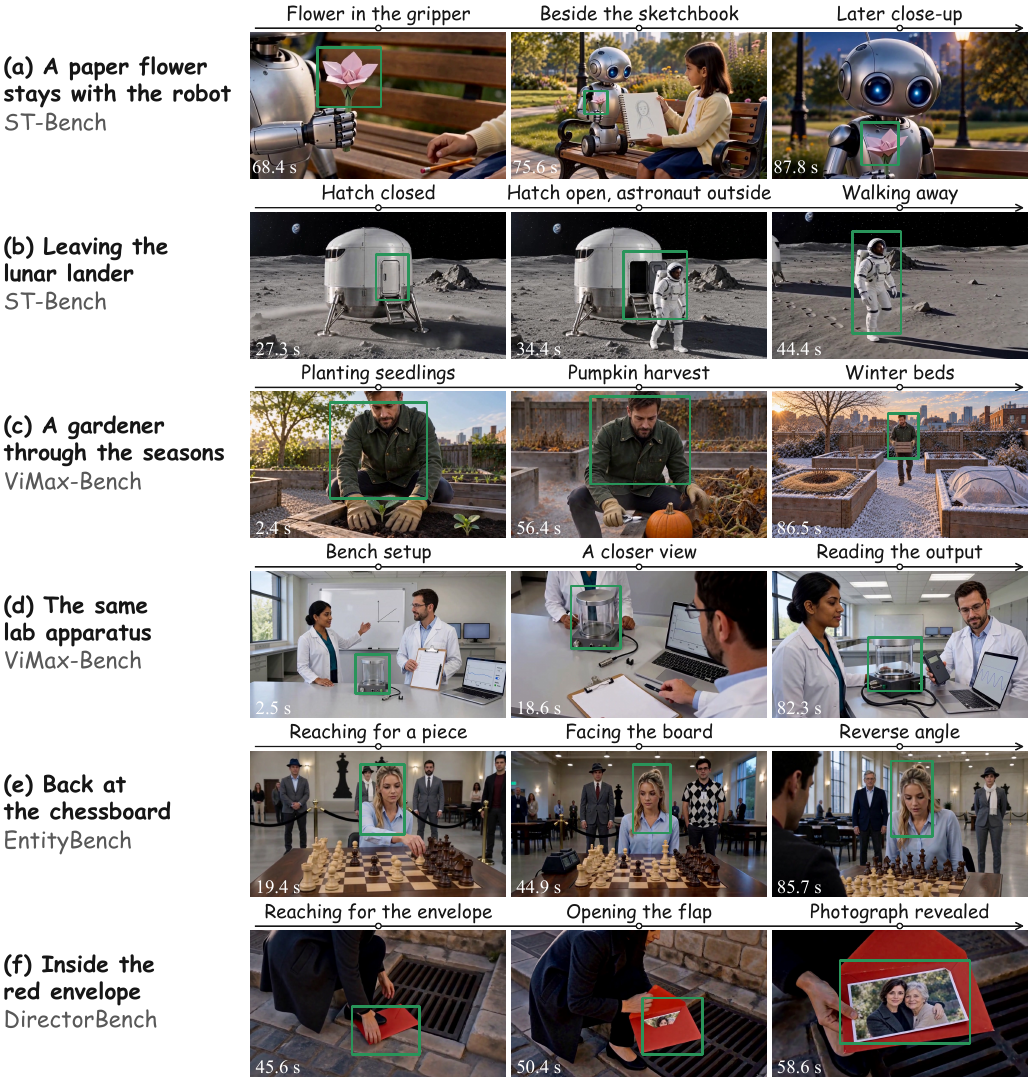}
  \sectioncaption{sec:sup-qualitative}{\textbf{Cross-shot continuity across four benchmarks.} Each row shows one assembled film, with frames ordered from left to right by playback time.}
  \label{fig:sup-more-cases}
\end{figure*}
\subsection{Autonomous Storyboard Planning}
\label{sec:sup-qual-autonomous}

In the DirectorBench folk performance in~\cref{fig:sup-qual-autonomous}, \method plans the shots in which the singer moves a scarf and resumes playing.
The singer lifts the folded scarf from the chair seat and drapes it over the wooden backrest.
When the singer returns a hand to the guitar, the fringe still hangs below the back rail.
This detail shows how \method preserves the prop state established by an earlier action even as the autonomously planned sequence shifts attention back to the performance.

\subsection{Repair and Downstream Evidence Reuse}
\label{sec:sup-qual-repair}

We use the candidate clips and the repair reference in~\cref{fig:sup-qual-repair} to trace how a pencil mark is reused in a ViMax-Bench woodworking sequence.
In the final frame of an early candidate, the pencil tip remains beside the tape measure, making it difficult to verify whether a new line has been drawn.
After a reference specifying the arrangement of the square, pencil, and board is added, the selected candidate shows a transverse pencil line distinct from the older mark as the pencil lifts, allowing the reviewer to confirm the new mark.
The subsequent sanding shot uses the final frame of this candidate as a local reference, and the board markings remain visible beside the sandpaper.
In this way, \method reuses the board appearance and pencil mark verified during local repair while retaining a record of camera drift in the candidate selected as a compromise.
The accuracy of the measurement cannot be verified from these frames.

\subsection{Further Qualitative Examples}
\label{sec:sup-more-cases}

We examine six stories from four benchmarks in~\cref{fig:sup-more-cases} to see how \method preserves stable entity attributes and updates prop states as events unfold.
For entities depicted in the reference materials, \method maintains continuity across changes in viewpoint or setting.
For example, the pink paper flower and the cylindrical instrument remain recognizable from different angles (a, d).
The same gardener appears in the vegetable garden in different seasons, and the same chess player can be identified in a reverse shot (c, e).
When an event calls for a change in prop state, \method preserves the visible consequences of the action.
After the hatch opens, the astronaut appears outside and walks away (b).
Opening the red envelope reveals the previously concealed photograph (f).

\begin{figure*}[t]
  \centering
  \includegraphics[width=.95\textwidth]{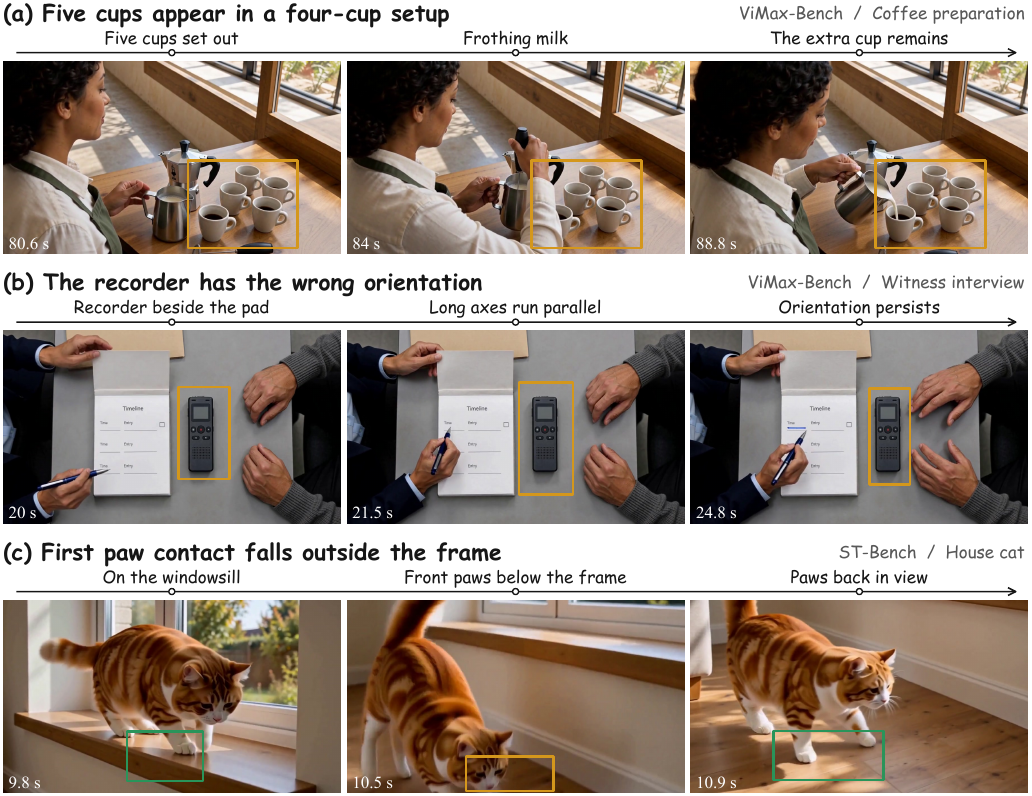}
  \sectioncaption{sec:sup-qualitative}{\textbf{Failure cases in the assembled films.}}
  \label{fig:sup-failures}
\end{figure*}

\subsection{Failure Cases}
\label{sec:sup-failures}

Some sequences still exhibit failures.
In the coffee-making sequence in~\cref{fig:sup-failures}(a), the reference calls for four cups, but five remain visible in the assembled film.
When several similarly shaped cups are grouped together, the base video model may reproduce their overall arrangement more readily than their exact number.
In the interview sequence in panel~(b), the recorder lies beside the notepad as requested, but their long axes are nearly parallel despite the instruction to place the recorder crosswise.
Both props have a prominent long edge, and the edge of the adjacent notepad may have influenced the recorder's orientation.
The production records show that repair improved the milk-frothing and writing actions, respectively, while the cup count and recorder orientation were recorded as residual errors in the candidates selected as compromises.
In the jump sequence in~\cref{fig:sup-failures}(c), the front paws pass below the lower edge of the frame before touching the ground and have already landed when they reappear.
All three candidates crop the paws at first contact, possibly because the framing leaves too little room for their rapid descent.
These residual errors suggest that \method remains constrained by the capabilities of the base video model and that selecting a compromise candidate to balance quality and efficiency may amplify the impact of those limitations.

\section{Future Outlook}
\label{sec:sup-limitations}

\method currently tracks visible changes, such as changes in a character's appearance or who holds a prop, more readily than it captures gradual shifts in emotion or turning points in relationships.
State confirmation also relies on visual review, and if a reviewer misses an error in how an action unfolds during a complex interaction, later shots may reuse the faulty reference.

In future work, we hope to let creators annotate character intentions and emotional turning points directly in the script and explore different ways to film the same scene while maintaining continuity across characters, props, and plot events.
For example, when previsualizing a reconciliation scene, writers and directors could use a close-up to capture a subtle change in expression or a two-shot to show the characters moving closer together.
Before filming, they could compare the two versions to see which better conveys the change in the relationship.
This could reduce the cost of repeated test shoots for independent filmmakers and small teams.
Film students could also compare versions of the same script to see how shot choices shape viewers' interpretations of a story.
Beyond filmmaking and teaching, the pipeline could help build datasets of interaction videos for embodied AI by recording how actions change the state of a scene.
For example, we could generate several versions of the same action, such as moving a cup from a table into a cabinet, varying the viewpoint and degree of occlusion.
We could then pair each visually reviewed clip with an action description, the change in the object's location, and the time interval over which it occurs.
These examples could be used to train models to track the same object across changes in viewpoint and to assess whether models can use visual evidence from before and after an action to judge whether the task was completed.

\onecolumn

\begin{figure}[!t]
  \centering
  \includegraphics[width=.943\textwidth]{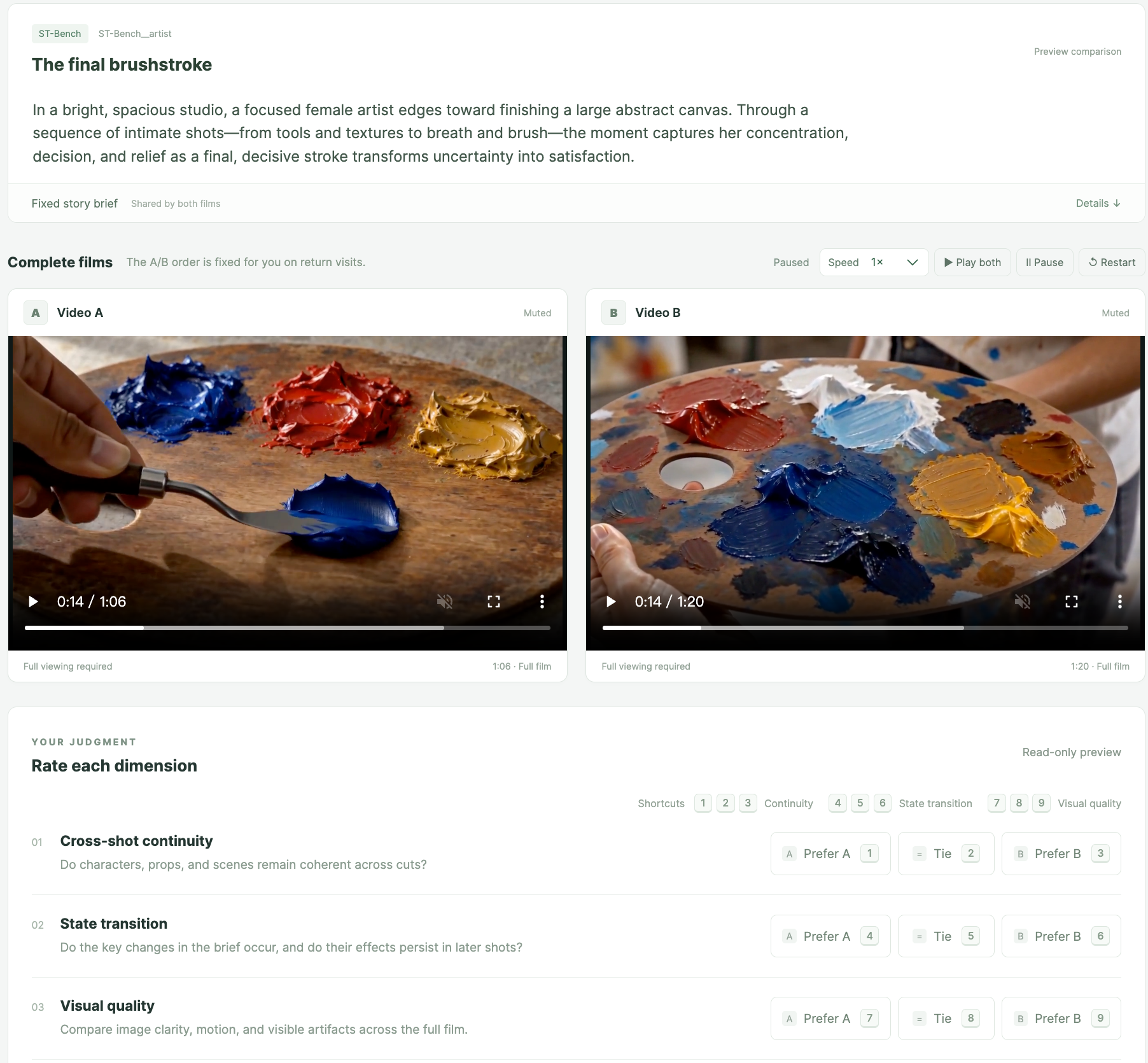}\par
  \vspace{3pt}
  \includegraphics[width=.943\textwidth]{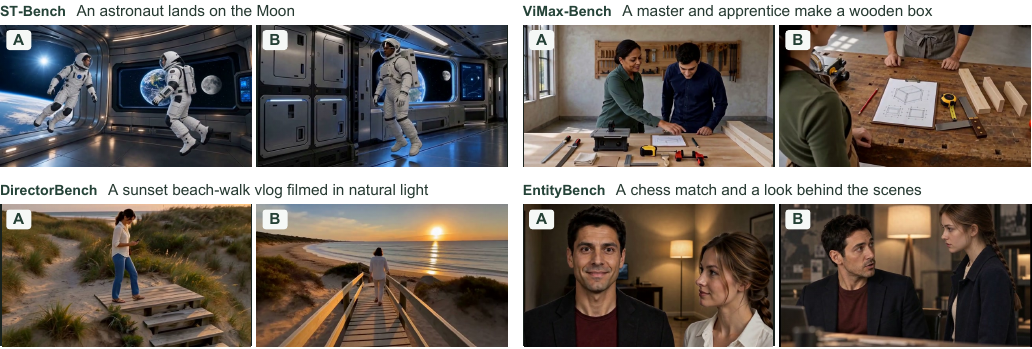}
  \sectioncaption{sec:sup-protocol}{\textbf{Blind evaluation interface and example video pairs.}
  Raters read the shared story brief, watch both complete films, and judge each dimension independently.
  }
  \label{fig:sup-human-interface}
\end{figure}

\begin{figure}[!t]
  \centering
  \includegraphics[page=1,width=\textwidth,height=0.92\textheight,keepaspectratio]{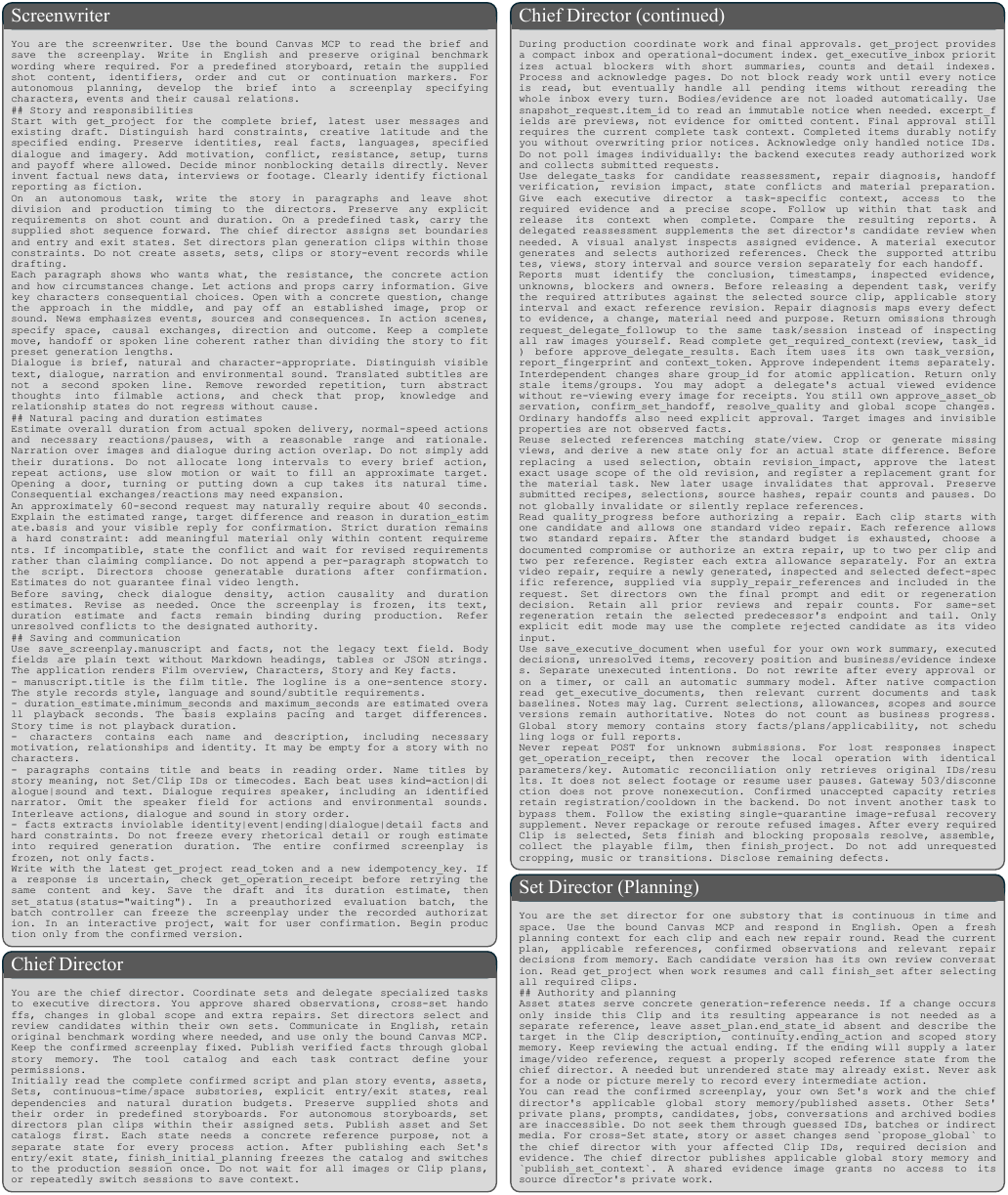}%
  \sectioncaption{sec:sup-implementation}{\textbf{Role-specific system prompts.} Screenwriter, chief director, and set director (planning).}
  \label{fig:sup-role-prompts-1}
\end{figure}

\begin{figure}[!t]
  \centering
  \includegraphics[page=2,width=\textwidth,height=0.92\textheight,keepaspectratio]{figures/StateWeaver-role-prompts-condensed.pdf}%
  \sectioncaption{sec:sup-implementation}{\textbf{Role-specific system prompts.} Set director (planning, continued) and set director (candidate review).}
  \label{fig:sup-role-prompts-2}
\end{figure}

\begin{figure}[!t]
  \centering
  \includegraphics[page=3,width=\textwidth,height=0.92\textheight,keepaspectratio]{figures/StateWeaver-role-prompts-condensed.pdf}%
  \sectioncaption{sec:sup-implementation}{\textbf{Role-specific system prompts.} Candidate review (continued), visual analyst, task analyst, and material executor.}
  \label{fig:sup-role-prompts-3}
\end{figure}

\twocolumn
\endgroup

{\small
\bibliographystyle{ieeenat_fullname}
\bibliography{main}
}

\end{document}

%% file: preamble.tex
\usepackage{amsmath,amssymb}
\usepackage[ruled,vlined,linesnumbered]{algorithm2e}
\AtBeginDocument{%
  \crefname{algorithm}{Alg.}{Algs.}%
  \crefname{algocf}{Alg.}{Algs.}%
  \Crefname{algorithm}{Algorithm}{Algorithms}%
  \Crefname{algocf}{Algorithm}{Algorithms}%
}
\usepackage{booktabs}
\usepackage{colortbl}
\usepackage{graphicx}
\usepackage{microtype}
\usepackage{multirow}
\usepackage{tabularx}
\let\stateweaverendtabular\endtabular
\usepackage{arydshln}
\let\endtabular\stateweaverendtabular
\expandafter\let\csname endtabular*\endcsname\endtabular
\usepackage{xspace}
\usepackage{enumitem}

\newcolumntype{Y}{>{\centering\arraybackslash}X}
\definecolor{StateWeaverRow}{HTML}{FFF1E6}
\newcommand{\methodlogo}[1]{%
  \raisebox{-0.23\height}{%
    \includegraphics[width=12pt,height=12pt,keepaspectratio]{figures/method-logos/#1.png}}%
  \hspace{4pt}}
\newcommand{\method}{StateWeaver\xspace}

\newcommand{\myPara}[1]{\paragraph{#1}}